\documentclass[lettersize,journal,onecolumn]{IEEEtran}
\usepackage{amsmath,amsfonts}

\usepackage{array}
\usepackage{caption}
\usepackage{textcomp}
\usepackage{stfloats}
\usepackage{url}
\usepackage{verbatim}
\usepackage{tabularx}
\usepackage[T1]{fontenc}
\usepackage{graphicx}
\usepackage{cite}
\usepackage{float}
\usepackage{longtable}
\usepackage{subfloat}
\usepackage{multirow}
\usepackage{soul}
\usepackage[table]{xcolor}
\usepackage{epstopdf}
\allowdisplaybreaks[0]
\usepackage{subfigure}
\usepackage[switch]{lineno}
\makeatletter
\newif\if@restonecol
\makeatother

\usepackage[linesnumbered,ruled,vlined]{algorithm2e}%[ruled,vlined]{
\usepackage{algpseudocode}
\usepackage{amsmath}
\usepackage{setspace}
\begin{document}
\title{Anti-Byzantine Attacks Enabled Vehicle Selection for Asynchronous Federated Learning in Vehicular Edge Computing}
\author{

{
	Cui Zhang, Xiao Xu, Qiong Wu*, Pingyi Fan, Qiang Fan, Huiling Zhu and Jiangzhou Wang

\thanks{

{	Cui Zhang is with the School of Internet of Things Engineering, Wuxi Institute of Technology, Wuxi 214121, China 

    Xiao Xu and Qiong Wu are with the School of Internet of Things Engineering, Jiangnan University, Wuxi 214122, China

	Pingyi Fan is with the Department of Electronic Engineering, Beijing National Research Center for Information Science and Technology, Tsinghua University, Beijing 100084, China 

	Qiang Fan is with Qualcomm, San Jose, CA 95110, USA 

	Jiangzhou Wang is with the School of Engineering, University of Kent, CT2 7NT Canterbury, U.K. 
   (* The corresponding author, email: qiongwu@jiangnan.edu.cn )

}
}
}
}
\maketitle

\begin{abstract}
In vehicle edge computing (VEC), asynchronous federated learning (AFL) is used, where the edge receives a local model and updates the global model, effectively reducing the global aggregation latency.Due to different amounts of local data,computing capabilities and locations of the vehicles, renewing the global model with same weight is inappropriate.The above factors will affect the local calculation time and upload time of the local model, and the vehicle may also be affected by Byzantine attacks, leading to the deterioration of the vehicle data. However, based on deep reinforcement learning (DRL), we can consider these factors comprehensively to eliminate vehicles with poor performance as much as possible and exclude vehicles that have suffered Byzantine attacks before AFL. At the same time, when aggregating AFL, we can focus on those vehicles with better performance to improve the accuracy and safety of the system. In this paper, we proposed a vehicle selection scheme based on DRL in VEC. In this scheme, vehicle’s mobility, channel conditions with temporal variations, computational resources with temporal variations, different data amount, transmission channel status of vehicles as well as Byzantine attacks were taken into account.Simulation results show that the proposed scheme effectively improves the safety and accuracy of the global model.

\end{abstract}

\section{Introduction}
\label{Introduction}
As vehicular networks advance, the Internet of Vehicle (IoV) emerges to enable some real-time applications like audio recognition and multimedia collaboration, aiming to enhance people's daily lives \cite{1,2}. For IoV, vehicles get information from environment and use their local information to train models in order to enhance vehicle service capabilities. Because the computational capability is finite\cite{46}, so vehicles have to request and send information to a cloud which has strong computational capability. Accordingly, the cloud will process the information and provide the relevant vehicles with computational results \cite{3}.
However, the cloud is often located at a distance from the vehicle, and will cause a long transmission delay\cite{50}. Vehicular edge computing (VEC) can solve this problem \cite{4}. In the VEC, the road side unit (RSU) installed alongside the road can be alternatively used as the edge so that it can decrease the latency \cite{5,52} .

Nonetheless, because of the concerns over data privacy, vehicles frequently exhibit hesitancy in sharing information with one another. As a result, the RSU might encounter difficulties in accumulating a sufficient amount of information to train a precise model. Federated learning (FL) offers a potential solution to address this issue effectively \cite{7}.
Specifically, in conventional federated learning, the RSU iteratively refines the global model over multiple rounds. First, the RSU initializes a global model. During every iteration, each vehicle initially gets the global model from the RSU, and trains the local model using local data. Before next round training, the vehicle uploads its local model to the RSU, and then the RSU aggregates all received local models to  renew the global model. The procedure continues until the global model converges or up to the maximum training round. In this way, it can eliminate privacy concerns by uploading model rather than the local information.

In traditional FL, the synchronous update of the global model is a requisite. That is to say, before updating the global model, the RSU is required to wait until each vehicle has submitted its local model in each round.
Nonetheless, some straggler vehicles probably have a high local training delay and thus move beyond the range of the RSU before completing the current round training of its local model. As a result, the RSU is unable to aggregate the local models of these straggler vehicles in time, that will affect the global model's precision. Asynchronous federated learning (AFL) offers a potential solution to address this issue effectively \cite{8}. Specifically, upon receiving an uploaded local model, the RSU will proceed to update the global model. The requirement of synchronous uploading of every local model can be avoided by this process \cite{9}.

In vehicular networks with edge assistance, vehicle mobility has to be taken into account \cite{10}. The variation of the vehicle position can lead to a change in distance from the RSU, thus causing a change in its model transmission time\cite{51}. Furthermore, the variation in data quantity and computation capabilities among vehicles will impact the local training time in AFL and time-varying features of channels will also affect the model upload rate.This is because channel conditions include path loss and channel gain, which in turn affect propagation rate, leading to changes in upload delay. 
Particularly, the time-changing mobility patterns of vehicles can result in varied transmission rates \cite{11,47}, consequently leading to diverse transmission delays, which refer to the time taken for uploading the local model \cite{12}. Moreover, the heterogeneous computing capabilities and data sizes of vehicles contribute to distinct local training delays for their respective local models.
In case that the duration between downloading  global model and uploading local model is large for a vehicle, RSU might have already revised the global model by using local models from other vehicles. Under this circumstance, this vehicle’s local model is staleness. Furthermorer, it will affect AFL’s training and also decrease global model’s precision \cite{13}. Thus, the vehicles’ mobility, data magnitude and computational resources are very important. It is crucial to give consideration to the vehicles’s mobility, data magnitude and computing capabilities to renew the global model with different weights and select the vehicles with better performance for global aggregation.

In AFL, vehicles’ data may be attacked by Byzantine attacks \cite{14}.
Specifically, the Byzantine attacks will affect the vehicles’ local data, tamper the data or labels and degrade the local models’ precision. In that case it can leads to global model’s reduction at the RSU. 
Thus, the prevention of Byzantine attacks is significantly important for AFL in VEC. Therefore, it is of great significance to consider vehicle’s mobility, time-changing computational capabilities, data size, time-changing channel properties, Byzantine attacks and select the proper vehicles to improve global model’s precision during VEC \cite{15}.

In this paper, by considering the factors above , we provide a method about vehicle identification and selection for AFL in VEC so that it can increase global model’s accuracy
\footnote{The source code has been released at: https://github.com/qiongwu86/By-AFLDDPG}.  
The main contributions of this paper are listed in the following:

\begin{itemize}
	\item[1)]In the process of selecting vehicles,we take vehicles with limited local data, poor computational capabilities and poor communication channel status due to interference into account, and employ deep reinforcement learning (DRL) for the selection of vehicle to engage in AFL, so that it can mitigate the influence on the global model’s aggregation.
	\item[2)]In the aggregation process of AFL,we take vehicle’s characteristics mobility, time-changing channel status and time-changing computational capabilities into account.Therefore,the appropriate local model of the vehicle will make a greater contribution to the global model so as to enhance global model’s precision.
	\item[3)]We consider the Byzantine attacks in AFL aggregation, and set threshold to identify and select vehicles to prevent the attack from affecting global model’s precision.
	\item[4)]The simulation outcomes indicate that the method we proposed is able to successfully enhance global model’s the security and precision .
\end{itemize}

The subsequent sections of this paper are structured as follows. 
Section II provides an overview of the relevant literature. 
Section III presents system model in detail. 
Section IV introduces DRL model. 
Section V introduces the proposed algorithm. 
Section VI shows the simulation outcomes. 
Section VII provides a summary of the key findings for this paper.

\section{RELATED WORK}
\label{RELATED WORK}
In vehicular networks, several studies have been conducted on federated learning.
In \cite{16}, Chu \emph{et al.} presented a distributed framework that relies on a highly effective federated DRL approach in order to manage the charging of plug-in electric vehicle platoons. It can ensure the scalability and privacy protection.
In \cite{17}, Liu \emph{et al.} proposed a FL based placement decision method of idle mobile charging stations to help the idle mobile charging stations to predict the future charging positions and thus can enhance the proportion of charged electric vehicles and reduce the charging expenses of them.
In \cite{18}, Liang \emph{et al.} introduced a semi-synchronized federated learning scheme which selected vehicles and aggregated dynamically to optimize the convergence speed and resource consumption of vehicles in IoV.
In \cite{19}, Kong \emph{et al.} presented  a cooperative vehicle positioning scheme that relies on FL, harnessing the capabilities of the social Internet of Things and collaborative edge computing. This innovative approach enables accurate positioning correction while also prioritizing user privacy protection.
In \cite{20}, Li \emph{et al.} proposed a federated multiagent DRL scheme for the distributed collaborative optimization for the selection of channels as well as power control in vehicle-to-vehicle communications.
In \cite{21}, Li \emph{et al.} provided a collaborative optimization scheme for vehicle selection and distribution of resources for FL.
In \cite{22}, Pokhrel \emph{et al.} introduced a privacy-preserving and communication-efficient federated learning scheme to enhance the performance of IoV. 
In \cite{23}, Zhou \emph{et al.} provided a two layer FL model in 6G supported vehicular networks with the aim of attaining heightened efficiency and precision while guaranteeing information privacy and minimizing communication costs.
In \cite{24}, Lv \emph{et al.} introduced a scheme using blockchain technology for FL to identify misbehavior. The scheme facilitates collaborative training using several distributed edge devices, simultaneously prioritizing information security and privacy in vehicular networks.
In \cite{25}, Zeng \emph{et al.} provided a FL framework enabled through wireless connectivity with large scale to design the autonomous controller for connected and autonomous vehicles. Nevertheless, these studies have not assessed the possibility that some straggler vehicles may exit RSU’s coverage ahead of uploading local models due to their mobility, which will affect global model’s precision.

There are many researches that have explored the concept of AFL in wireless networks.
In \cite{26}, Pan \emph{et al.} introduced a method which used deep Q-learning and AFL to optimize throughput, taking into account the enduring requirements of ultra-reliable and communication with low delay .
In \cite{27}, Bedda \emph{et al.} proposed an asynchronously weight updating FL scheme which had low-latency and high-efficiency to allocate resources for different users.
In \cite{28}, Liu \emph{et al.} proposed an AFL arbitration framework utilizing bidirectional long short-term memory (LSTM) and incorporating an attention mechanism to get higher accuracy of model and reduced overall communication rounds.
In \cite{29}, Wang \emph{et al.} proposed an AFL scheme which included a novel centralized fusion algorithm to enhance the training efficiency.
In \cite{30}, Lee \emph{et al.} proposed an adaptive scheduling strategy using AFL in the wireless distributed learning network. Nevertheless, the studies above have not taken the vehicles’ characteristics including vehicle mobility, data size and computational capability into account to choose vehicles for AFL in vehicular networks.
In \cite{44}, Wu \emph{et al.} proposed a scheme for selecting vehicles to participate in AFL based on deep reinforcement learning,however, this paper has no protective measures for security, which can lead to significant fluctuations in the accuracy of the global model. This is a risk that cannot be ignored in the context of connected vehicles.

There are some works studying the Byzantine attack in wireless networks.
In \cite{31}, Vedant \emph{et al.} provided a method that took practical Byzantine fault tolerance into account to secure the energy exchanging between electric vehicles and the grid and thus enhanced the privacy and security. 
In \cite{32}, Ma \emph{et al.} provided a mechanism based on credibility and an efficient protocol which can protect privacy for gradient aggregation to combat Byzantine attacks within Non-IID(Non-Independent and identically distributed) datasets which are usually gathered from heterogeneous ships.
In \cite{33}, Sheikh \emph{et al.} focused on the energy exchanging procedure between electric vehicles and power grid in a blockchain consensus system that employs Byzantine as its underlying mechanism to improve the security of the system.
In \cite{34}, Wang \emph{et al.} introduced a decentralized system for collecting traffic data securely and protecting privacy using blockchain technology, while using the capabilities of fog/edge computing infrastructure.
In \cite{35}, Huang \emph{et al.} proposed a scheme for integrating vehicle spacing data and calculating the average spacing within a defined area over a brief timeframe to estimate traffic density and thus can resist Byzantine attack.
In \cite{36}, Chen \emph{et al.} provided a novel Byzantine fault tolerant FL scheme in terms of vehicles and each vehicle used the publicly-verifiable secret sharing mechanism to safeguard the confidentiality of the model.
In \cite{37}, Wang \emph{et al.} introduced a decentralized system for collecting traffic data securely and protecting privacy on the blockchain, while employing DL techniques to mitigate Byzantine and sybil attacks.
In \cite{38}, Xu \emph{et al.} provided a networking approach resilient to Byzantine faults, along with two strategies of resource allocating for fog computation enabled internet of things.
In \cite{39}, Wang \emph{et al.} provided a method of integrating vehicle spacing data for estimating average spacing. They also proposed different estimation methods for different kinds of Byzantine attacks. 
However, they have not considered the AFL in vehicular networks. 
In \cite{45}, Fang \emph{et al.}provided a scheme to prevent Byzantine attacks in AFL, however, compared to our scheme,this scheme only performs one round of filtering in the global aggregation stage, which will result in the attacked vehicles also participating in AFL training, thereby affecting the overall training efficiency.In addition,this scheme has not comprehensively consider the characteristics of each vehicles in the aggregation process,which can reduce the impact of some bad clients.our plan has taken Byzantine attack into consideration in DDPG in advance, and use the loss of the model instead of size and amplitude as the standard, which reduces computational complexity and has better generalization ability.
%到这

Based on our current understanding, there is no study jointly taking vehicles’ characteristics  and Byzantine attacks into account to propose vehicle selection scheme for the AFL in VEC, which serves as our motivation to carry out this work.

\section{SYSTME MODEL}
\label{SYSTME MODEL}
\begin{figure}
	\center
	\includegraphics[scale=0.9]{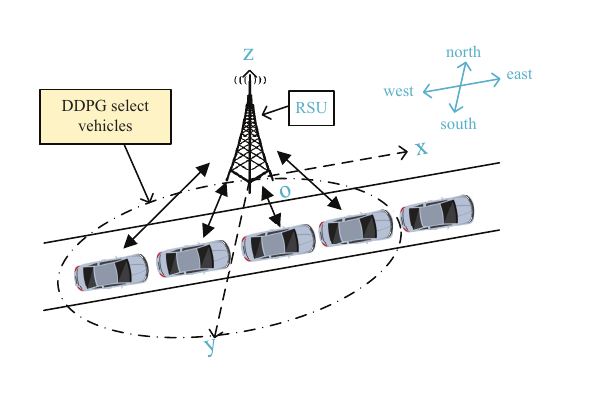}
	\caption{System model}
	\label{fig1}
\end{figure}
The system model is illustrated in Fig. \ref{fig1}. We consider a scenario which consists of a RSU and vehicles whose number is $K$. The vehicles are driving in the coverage area of the RSU.
We set the bottom of the RSU as the origin of the coordinate system. Specifically, the $x$-axis is oriented towards the east, the $y$-axis is oriented towards the south, and the $z$-axis is orthogonal to both $x$-axis and $y$-axis, aligned with the antenna of RSU direction.  All vehicles move towards east with the same velocity $v$, and the duration is subdivided into separate time slots. Once the vehicle enters the coverage range of the local RSU, it will send a request message to the local RSU,which can match most of cases in the highway scenarios. For simplicity, we assume that there are no vehicles entering or exiting the coverage area of the local RSU in one round in the discussed scenarios, so the number of vehicles in the local RSU is fixed in one round. Different vehicles  have different amounts of data $D_n$ and have different computing capabilities, where the amount of data for vehicle $n\left( {1 \le n \le K} \right)$ is denoted by $D_n$. Also, vehicle mobility leads to the time-varying channel condition.

Here, a local model may be affected by external interference and Byzantine attacks.  We assume that the RSU has a clean dataset which means that this dataset will not be influenced by any interference or attack, that is, the trusted dataset $D_{RSU}$.

\section{Deep Reinforcement Learning Model}
As some vehicles have poor performance such as the limited quantity of local information and degraded local model affected by external interference, we employ DRL to solve the problem of initial vehicle selection in our AFL\cite{40,48,53,54}. Specifically, the process has $N$ time slots in each episode. In one time slot, the state, action and reward function of the system are constructed first, and then the system makes a corresponding action according to the current state, where the corresponding reward is generated. Then the system reaches the next state and begins the next time slot. Next, we will define system state, action, and reward function respectively.
\subsection{State}
For each vehicle, the delay of the local training is related to the data's quantity and computational capacity of the vehicle, the uploading delay of local model is related to channel condition and transmission rate. Therefore, the system state under time slot $t$ can be characterized as
\begin{equation}
	s\left( t \right) = \left( {R\left( t \right),\mu \left( t \right),{d_x}\left( t \right),a\left( {t - 1} \right)} \right),
	\label{eq1}
\end{equation}
where $R\left( t \right)$ is all vehicles' transmission rate during $t$, which can be characterized as $R\left( t \right) = \left( {{R_1}\left( t \right),{R_2}\left( t \right), \ldots {R_K}\left( t \right)} \right)$, 
$\mu \left( t \right)$ represents each vehicle's computational resources  during $t$, which can be characterized as $\mu \left( t \right) = \left( {{\mu _1}\left( t \right),{\mu _2}\left( t \right), \ldots {\mu _K}\left( t \right)} \right)$.In our simulation, it follows a truncated Gaussian distribution, different vehicles have different computing resources. Meanwhile, due to the fact that the computing resources of vehicles are not affected by external conditions and only undergo slight changes due to the influence of the vehicles themselves, in order to simulate a real connected vehicle environment, at each time slot $t$, the computing resources of each vehicle will be reallocated, which also conforms to a truncated Gaussian distribution, which is consistent with the situation of most kinds of vehicles.
${d_x}\left( t \right)$ represents set of $x$-axis coordinates of all vehicles during $t$, which can be characterized as ${d_x}\left( t \right) = \left( {{d_{1x}}\left( t \right),{d_{2x}}\left( t \right), {d_{3x}}\left( t \right),\ldots {d_{Kx}}\left( t \right)} \right)$,
the last is $a\left( {t - 1} \right)$, which represents the action during $t-1$.

Next we will present $R\left( t \right)$ and ${d_x}\left( t \right)$.

${P_n}\left( t \right)$ represents the vehicle$n$'s position during $t$. Then ${P_n}\left( t \right)$ can be denoted as $\left( {{d_{nx}}\left( t \right),{d_y},0} \right)$, where ${d_{nx}}\left( t \right)$ and ${d_y}$ are the $x$-axis and $y$-axis coordinates, respectively, of vehicle $n$ at time slot $t$.
The $d_y$ above is an invariant and then we can get ${d_{nx}}\left( t \right)$:
\begin{equation}
	{d_{nx}}\left( t \right) = {d_{n0}} + vt,
	\label{eq2}
\end{equation}
The ${d_{n0}}$ above represents vehicle's starting location  $n$ alongside the $x$-axis.

The antenna's height is established as $H_R$, in addition, the position of the RSU's antenna is set as ${P_R} = \left( {0,0,{H_R}} \right)$.
Subsequently, the distance separating vehicle from RSU's antenna during $t$ is
\begin{equation}
	{d_n}\left( t \right) = \left\| {{P_n}\left( t \right) - {P_R}} \right\|.
	\label{eq3}
\end{equation}

Assume that vehicle$n$'s transmission rate during $t$ is ${R_n}\left( t \right)$.
According to the Shannon's theorem, ${R_n}\left( t \right)$ can be expressed as
\begin{equation}
	{R_n}\left( t \right) = B{\log _2}\left( {1 + {{{p_0}{\rm{\cdot}}{h_n}\left( t \right){\rm{\cdot}}{{\left( {{d_n}\left( t \right)} \right)}^{ - \alpha }}} \over {{\sigma ^2}}}} \right).
	\label{eq4}
\end{equation}
The $B$ above represents bandwidth, ${p_0}$ represents vehicles' transmission power which is a fixed value, the channel gain during $t$ can be defined as ${h_n}\left( t \right)$, and the path loss exponent can be defined as $\alpha $ , and ${\sigma ^2}$ is the power of noise.

The auto-regressive model is employed so that we can describe  ${h_n}\left( t \right)$ and ${h_n}\left( {t - 1} \right)$ as
\cite{41}:
\begin{equation}
	{h_n}\left( t \right) = {\rho _n}{h_n}\left( {t - 1} \right) + e\left( t \right)\sqrt {1 - \rho _n^2}.
	\label{eq5}
\end{equation}
The ${{\rho }_{n}}$ above represents the correlation coefficient of normalized channels across sequential time slot. Meanwhile, $e\left( t \right)$ represents an error vector which conforms to complex Gaussian distribution, what's more, it also associate with ${{h}_{n}}\left( t \right)$.
 ${{\rho }_{n}}={{J}_{0}}\left( 2\pi f_{d}^{n}t \right)$ based on Jake's fading spectrum, where ${{J}_{0}}\left( \text{ }\!\!\cdot\!\!\text{ } \right)$ is the zeroth-order Bessel function of the first kind and $f_{d}^{n}$ is the Doppler frequency of vehicle $n$ with
\begin{equation}
	f_d^n{\rm{ = }}{v \over \Lambda_0 }\cos \theta,
	\label{eq6}
\end{equation}
where $\Lambda_0$ represents wavelength, and $\theta$ represents angle formed by the orientation of motion ${{x}_{0}}=\left( 1,0,0 \right)$ and uplink communication orientation ${{P}_{R}}-{{P}_{n}}\left( t \right)$ satisfying
\begin{equation}
	\cos \theta  = {{{x_0}{\rm{\cdot}}\left( {{P_R} - {P_n}\left( t \right)} \right)} \over {\left\| {{P_R} - {P_n}\left( t \right)} \right\|}}.
	\label{eq7}
\end{equation}

\subsection{Action}
We aim to identify and select vehicles for AFL, where the action of the system during $t$ can be denoted as
\begin{equation}
	a\left( t \right) = \left( {{\lambda _1}\left( t \right),{\lambda _2}\left( t \right), \ldots ,{\lambda _K}\left( t \right)} \right).
	\label{eq8}
\end{equation}
The ${\lambda _n}\left( t \right),n \in \left[ {1,K} \right]$ above, represents the probability of selecting vehicle $n$ that can be initialized as  ${\lambda}\left( 0 \right) = 1$.
At the same time, in order to prevent vehicles that receive Byzantine attacks from joining AFL and get the selected vehicles, we define an another variable ${a_d}\left( t \right)$, which can be described as $ \left( {{a_{d1}}\left( t \right),{a_{d2}}\left( t \right), \ldots, {a_{dK}}\left( t \right)} \right)$.We first check for Byzantine attacks before selecting the selected vehicle. This part is detailed in subsection 5.3, and only the threshold is not exceeded and ${\lambda _n}\left( t \right)$ is no less than 0.5, the corresponding ${a_{dn}}\left( t \right)$ is set as 1 and the vehicle will be selected; otherwise  ${a_{dn}}\left( t \right)$ is set as 0. In this case, ${a_d}\left( t \right)$ consists of binary variables. The reason for why adopted the DDPG (Deep Deterministic Policy Gradient) algorithm in the DRL stage is that DDPG is suitable for continuous action state spaces.Other algorithms applicable to the continuous action space, such as the DPG(Deterministic Policy Gradient),TD3 ( Delay DDPG),compared to DDPG, the DPG algorithm usually uses Monte Carlo estimated action value functions when updating policies, which may result in large update variances and affect the convergence and stability of the algorithm. TD3 uses two Q-networks to reduce overestimation problems, but managing the training and updating of the two Q-networks also increases the complexity and computational cost of the algorithm.Therefore, considering the advantages and disadvantages of these algorithms, we chose DDPG.Therefore, the action is set to a continuous value of 0-1. The reasons for our adoption of DDPG are explained in detail in subsection 5.1.

\subsection{Reward}
By considering global model's latency and precision, the reward during $t$ can be described as
\begin{equation}
	\scalebox{0.8}{$
	\begin{split}
		&r\left( t \right) =\\  &- {K \over {\sum\limits_{n = 1}^K {{\lambda _n}\left( t \right)} }}\left[ {{W_1}Loss\left( t \right) + {W_2}{{\sum\limits_{n = 1}^K {\left( {T_l^n + T_u^n\left( t \right)} \right){a_{dn}}\left( t \right)} } \over {\sum\limits_{n = 1}^K {{a_{dn}}\left( t \right)} }}} \right],
		\label{eq9}
	\end{split}
	$}
\end{equation}
where ${W_1}$ and ${W_2}$ are the non-negative weight factors, $Loss\left( t \right)$ is the loss that gets from the AFL training which we will introduce later.As the average probability of vehicles being selected, $K \over {\sum\limits_{n = 1}^K {{\lambda _n}\left( t \right)} }$is used as a weight to reduce the impact when only a few vehicles are selected, as the inherent characteristics of individual vehicles may cause significant error fluctuations in this situation.
Next we will introduce local training dalay $T_l^n$ and transmission delay $T_u^n\left( t \right)$.

Vehicle $n$  employs its own information for training local model, then the corresponding training delay $T_l^n$ is expressed as
\begin{equation}
	T_l^n = {{{D_n}{C_0}} \over {{\mu _n}}},
	\label{eq10}
\end{equation}
where ${C_0}$ is the CPU cycles, ${\mu _n}$ is the computing resources of vehicle $n$ and is measured with CPU cycles frequency.

$T_u^n\left( t \right)$ represents transmission latency of uploading local model during $t$, which can be defined as
\begin{equation}
	T_u^n\left( t \right) = {{\left| w \right|} \over {{R_n}\left( t \right)}}.
	\label{eq11}
\end{equation}
The $\left| w \right|$ above represents each vehicle's local model's size .

Then system's expected long-term discounted reward can be expressed as
\begin{equation}
	J\left( \mu  \right) = E\left[ {\sum\limits_{t = 1}^N {{\gamma ^{t - 1}}r\left( t \right)} } \right].
	\label{eq12}
\end{equation}
The $\gamma  \in \left( {0,1} \right)$ above represents discounted factor, $N$ represents time slots' overall quantity, and $\mu $ represents system's policy. Our goal is to identify the most effective strategy to maximize system's expected long-term discount reward.

\section{Enhanced Vehicle Selection for AFL}
In this section, we will introduce the system framework. We first introduce the vehicle selection for AFL based on DRL\cite{49}, and then we proposed an enhanced mode to further select vehicles to prevent the Byzantine attack.

\subsection{Training stage}
We employ the deep deterministic policy gradient (DDPG) algorithm\cite{43} to solve the problem of DRL.It is based on the DQN (Deep Q-Network) algorithm and uses deep neural networks to approximate the Q-value function and policy function, thereby achieving prediction and optimization of continuous actions. Unlike DQN, DDPG has a stronger problem-solving ability\cite{55}.It solves the problem of actors criticizing the correlation between neural networks before and after parameter updates, note that DQN cannot be used for continuous actions, but DDPG can be used to solve reinforcement learning problems in the continuous state space and continuous action space. Considering that the state and action spaces are continuous in our discussed scenarios about the vehicular communications and computing, DDPG is a promising selection, thus we adopt DDPG in this work.

DDPG is established based on the actor-critic network. The actor network and critic network are all constructed as the fully connected deep neural networks (DNN).
Actor network is adopted to approximate the system policy with estimated policy ${\mu _\delta }$.
After observing the state based on ${\mu _\delta }$, it can then output the corresponding action while the critic network is adopted to evaluate the policy selected by actor network.

In addition, DDPG also includes a target network, which consists of both target actor and critic network, constructed with identical architectures to actor and critic network.

The parameters of actor network and critic network are $\delta$ and $\xi$, respectively, while their optimized parameters are ${\delta ^{'}}$ and ${\xi ^ {'}}$, respectively. The parameter of target actor network is ${\delta _1}$, and the parameter of target critic network is ${\xi _1}$. The target network's parameter which needs to update is $\tau $, and the noise during $t$ is ${\Delta _t}$, and the scale of mini-batch is $I$.
The pseudocode of the algorithm is shown in Algorithm \ref{al1}.

In Algorithm \ref{al1}, RSU first initialize $\delta $ and $\xi $ randomly, ${\delta _1}$ and ${\xi _1}$ are initialized as $\delta $ and $\xi $, respectively.
${R_B}$ is also initialized as an empty set.

The algorithm will be executed for $N$ time slots. In each time slot $t$, the actor network outputs ${\mu _\delta }\left( {s|\delta } \right)$ according to the current state, action contains random noise added by RSU so that we can calculate $a\left( t \right) = {\mu _\delta }\left( {s\left( t \right)|\delta } \right) + {\Delta _t}$.
Then ${a_d}\left( t \right)$ is used to select $K_t$ vehicles and all selected vehicles will participate in the training of AFL.
\begin{algorithm}[htbp]
	\caption{The Training Stage in DDAFL system}
	\label{al1}
	\KwData{$\gamma$, $\tau$, $\delta$, $\xi$, $a\left( 0 \right)$}
	\KwResult{optimal ${\delta }^{'}$, $\xi^{'}$}
	Set $\delta$, $\xi$\ in random\;
    Set $a\left( 0 \right)=\left( 1,1,\ldots ,1 \right)$\;
	Set target network by ${{\delta }_{1}}\leftarrow \delta$, ${{\xi }_{1}}\leftarrow \xi$\;
	Set $R_B$ as an empty set\;
	\While{$episode\leq E_{m}$ }
	{
		Reinitialize the simulation settings and RSU's global model.\;
		Obtain the first $s\left( 1 \right)$\;
		\While{$t \leq N$ }
		{
			Create a course of action based on the existing policy and incorporate exploratory noise. ie., $a={{\mu }_{\delta }}\left( s|\delta  \right)+{{\Delta }_{t}}$ \;
			Calculate the value of $a_d$, and identify the vehicles that will be chosen\;
			The vehicles that have been choosen perform AFL depending on the weight so that it can update global model.\;
			Obtain the reward $r$ alongwith $s'$\;
			Put the tuple $\left( s,a,r,s' \right)$ into $R_B$\;
			\While {tuples' quantity in $R_B > I$}
			{
				Chose $I$ tuples in $R_B$ to form a mini-batch \;
				Reducing the loss in order to revise critic network in accordance with Eq. \eqref{eq16}\;
				Revise the actor network in accordance with Eq. \eqref{eq17}\;
				Revise target networks in accordance with Eqs. \eqref{eq18} and \eqref{eq19}.
			}
		}
	}
\end{algorithm}

Next, we will introduce the training of AFL, which is the step 10.

The vehicle ${V_k}$ will train local model based on CNN through $l$ rounds. Note that for time slot $1$, depending on convolutional neural network (CNN), global model ${w_0}$ is initialized with random value by RSU, then each vehicle downloads the global model ${w_{0}}$ from the RSU and then trains the local model.

For each round $j$, the input of the CNN is the local data of $V_k$ and the output is the $\hat{y_{a}}$,which is data's label ${y_a}$ $(a=1,2,...,D_k)$'s predicted result. Then the loss of the local model updated in the previous round ${w_{k,j-1}}$ is calculated according to the cross-entropy loss function, i.e.,

\begin{equation}
	{{f}_{k}}\left( {{w}_{k,j-1}} \right)\text{=}-\sum\limits_{a=1}^{{{D}_{k}}}{{{y}_{a}}\log }{\hat{y_{a}}}\,.
	\label{eq20}
\end{equation}
Note that for round $1$, the local model is the downloaded global model ${w_{t-1}}$.

The local model of vehicle $V_k$ employs stochastic gradient descent (SGD) algorithm to perform update, i.e.,
\begin{equation}
	{w_{k,j}} = {w_{k,j-1}} - \eta \nabla {f_k}\left( {{w_{k,j-1}}} \right).
	\label{eq21}
\end{equation}
The $\nabla {f_k}\left( {{w_{k,j-1}}} \right)$ above represents ${f_k}\left( {{w_{k,j-1}}} \right)$'s gradient  , and $\eta $ above represents learning rate of this algorithm.
After that, the training enters the next round and keeps updating local model until the rounds' number is $L$, then outputs the local model ${w_k}$ is updated.
After that vehicle $V_k$ calculates ${f_k}\left( {{w_k}} \right)$ according to Eq. \eqref{eq20}.

%The loss of ${w_k}$ can be expressed as
%\begin{equation}
%{f_k}\left( {{w_k}} \right){\rm{ = }} - \sum\limits_{a = 1}^{{D_i}} {{y_a}%\label{eq22}
	%\end{equation}
	
	For the traditional AFL, once a vehicle finishes training its local model, it transfers local model to the RSU for the purpose of updating global model. Then the vehicle obtains the global model for the purpose of training local model again. However, the local trainging delay and transmission delay may be large for the vehicle, thus when the vehicle is training or uploading its local model, other vehicles may have uploaded their local models to the RSU. In this case, the RSU would use other vehicles' local models to update the global model. In addition, the local model maybe stale because of the considerable latency in local training  and transmission. Hence, we consider the effect of the latency in local training and transmission to calculate weighted local model and each selected vehicel ${V_k}$ uploads its weighted local model once its weighted local model is obtained.
	
	Specifically, vehicle $V_k$ calculates the local training delay weight as
	\begin{equation}
		{\beta _{1,k}} = {M_1}^{T_l^{{V_k}} - 0.5},
		\label{eq23}
	\end{equation}
	where ${M_1} \in \left( {0,1} \right)$ is a parameter that can make ${\beta _{1,k}}$ decrease when the local training delay increases. $T_l^{{V_k}}$ above represents vehicle ${V_k}$'s latency in local training, which is calculated according to Eq. \eqref{eq10}.
	
	Since the downloading time of the global model can be neglected compared with the uploading time of the local model, ${V_k}$ calculates the transmission delay weight as
	\begin{equation}
		{\beta _{2,k}}\left( t \right) = {M_2}^{T_u^{{V_k}}\left( t \right) - 0.5},
		\label{eq24}
	\end{equation}
	where ${M_2} \in \left( {0,1} \right)$ can make ${\beta _{2,k}}$ decrease when the transmission delay increases. $T_u^{{V_k}}\left( t \right)$ above represents vehicle ${V_k}$'s transmission latency, and it is calculated according to Eq. \eqref{eq11}.
	
	Then ${V_k}$ calculates the weighted local model as
	\begin{equation}
		{w_{kw}} = {\beta _{1,k}}\cdot {w_k}\cdot {\beta _{2,k}}.
		\label{eq25}
	\end{equation}
	
	Once the above weighted local model is calculated, ${V_k}$ provides RSU with $w_{kw}$, ${f_k}\left( {{w_k}} \right)$, $T_u^{{V_k}}\left( t \right)$ and $T_l^{{V_k}}$, and global model is updated by the RSU as
	\begin{equation}
		{w_{new}} = \beta {w_{old}} + \left( {1 - \beta } \right){w_{kw}},
		\label{eq26}
	\end{equation}
	where ${w_{old}}$ above represents the present global model, ${w_{new}}$ represents global model which has been updated and then $\beta  \in \left( {0,1} \right)$ represents aggregation ratio. Note that upon the RSU receiving the initial local model with weight in time slot $t$, ${w_{old}} = {w_{t - 1}}$. Particularly, when the RSU receives the first weighted local model in time slot 1, ${w_{old}} = {w_{0}}$. After that the RSU sends back the global model ${w_{new}}$ to $V_k$.
	
	When the training of AFL in time slot $t$ is near end, the RSU obtains the weighted local model from the last vehicle of the selected vehicles, thus the RSU gets the updated global model ${w_t}$ and sends back ${w_t}$ to the last vehicle.
	
	After that, the RSU calculates the average loss of all vehicles' local models as
	\begin{equation}
		Loss\left( t \right) = {1 \over {{K_l}}}\sum\limits_{k = 1}^{{K_l}} {{f_k}\left( {{w_k}} \right)}.
		\label{eq27}
	\end{equation}
	Here step 10 has been completed.
	
	Then RSU calculates reward of $t$ according to Eq. \eqref{eq9}. Afterwards, every vehicle within the RSU's coverage calculates its new position and transmission rates according to Eq. \eqref{eq2} and Eq. \eqref{eq4} and get the computing resources according to the truncated Gaussian distribution, and sends them to the RSU,  based on which RSU can get $s\left( t+1 \right)$.
	
	After that, the RSU stores the tuple $\left( {s\left( t \right),a\left( t \right),r\left( t \right),s\left( t+1 \right)} \right)$ into ${R_B}$. At this time, the algorithm inputs the next state ${s\left( t+1 \right)}$ directly to actor network, then it goes to the following round in cases where tuples' quantity in ${R_B}$ is less than $I$.  
	
	When tuples' number in ${R_B}$ is larger than $I$, RSU will select $I$ tuples from ${R_B}$ to compose a mini-batch in random.
	Let $\left( {{s}_{i}},{{a}_{i}},{{r}_{i}},s_{i}^{'} \right)$ be tuple $i$ in the mini-batch, where $i\in \left[ 1,2,\ldots ,I \right]$.
	
	We set the action value function as ${Q_{{\mu _\delta }}}\left( {s\left( t \right),a\left( t \right)} \right)$ which can be described as follows
	\begin{equation}
		{Q_{{\mu _\delta }}}\left( {s\left( t \right),a\left( t \right)} \right) = {E_{{\mu _\delta }}}\left[ {\sum\limits_{{k_1} = t}^N {{\gamma ^{{k_1} - t}}r\left( {{k_1}} \right)} } \right],
		\label{eq13}
	\end{equation}
	which is the expected long term discounted reward of the system during $t$.
	
	Previous studies have demonstrated that ${\nabla _\delta }J\left( {{\mu _\delta }} \right)$ can be replaced by ${\nabla _\delta }{Q_{{\mu _\delta }}}\left( {s\left( t \right),a\left( t \right)} \right)$ \cite{40}.
	We cannot use Bellman equation to solve ${Q_{{\mu _\delta }}}\left( {s\left( t \right),a\left( t \right)} \right)$ because of the space of action $a(t)$ is continuous.
	In order to address this issue, the ${Q_{{\mu _\delta }}}\left( {s\left( t \right),a\left( t \right)} \right)$ is approximated with ${Q_\xi }\left( {s\left( t \right),a\left( t \right)} \right)$ through critic network's ratio $\xi $.
	
	Next input $s_{i}^{'}$ into the target actor network, and the target actor network will get the output $a_{i}^{'}={{\mu }_{{{\delta }_{1}}}}\left( s_{i}^{'}|{{\delta }_{1}} \right)$.
	Then $s_{i}^{'}$ and $a_{i}^{'}$ act as the input of the target critic network, the output is action value function ${{Q}_{{{\xi }_{1}}}}\left( s_{i}^{'},a_{i}^{'} \right)$.
	Then the target value of tuple $i$ can be calculated as
	\begin{equation}
		{{y}_{i}}={{r}_{i}}+\gamma {{Q}_{{{\xi }_{1}}}}\left( s_{i}^{'},a_{i}^{'} \right){{|}_{a_{i}^{'}={{\mu }_{{{\delta }_{1}}}}\left( s_{i}^{'}|{{\delta }_{1}} \right)}}.
		\label{eq14}
	\end{equation}
	
	After that put ${s_i}$ and ${a_i}$ to critic network, and we can get ${Q_\xi }\left( {{s_i},{a_i}} \right)$.
	Then the loss of tuple $i$ is calculated as
	\begin{equation}
		{L_i} = {\left[ {{y_i} - {Q_\xi }\left( {{s_i},{a_i}} \right)} \right]^2}.
		\label{eq15}
	\end{equation}
	
	When all the tuples are put into the networks, RSU can calculate the loss function as
	\begin{equation}
		L\left( \xi  \right) = {1 \over I}\sum\limits_{i = 1}^I {{L_i}}.
		\label{eq16}
	\end{equation}

\begin{figure*}
	\centering
	\includegraphics[width=16cm]{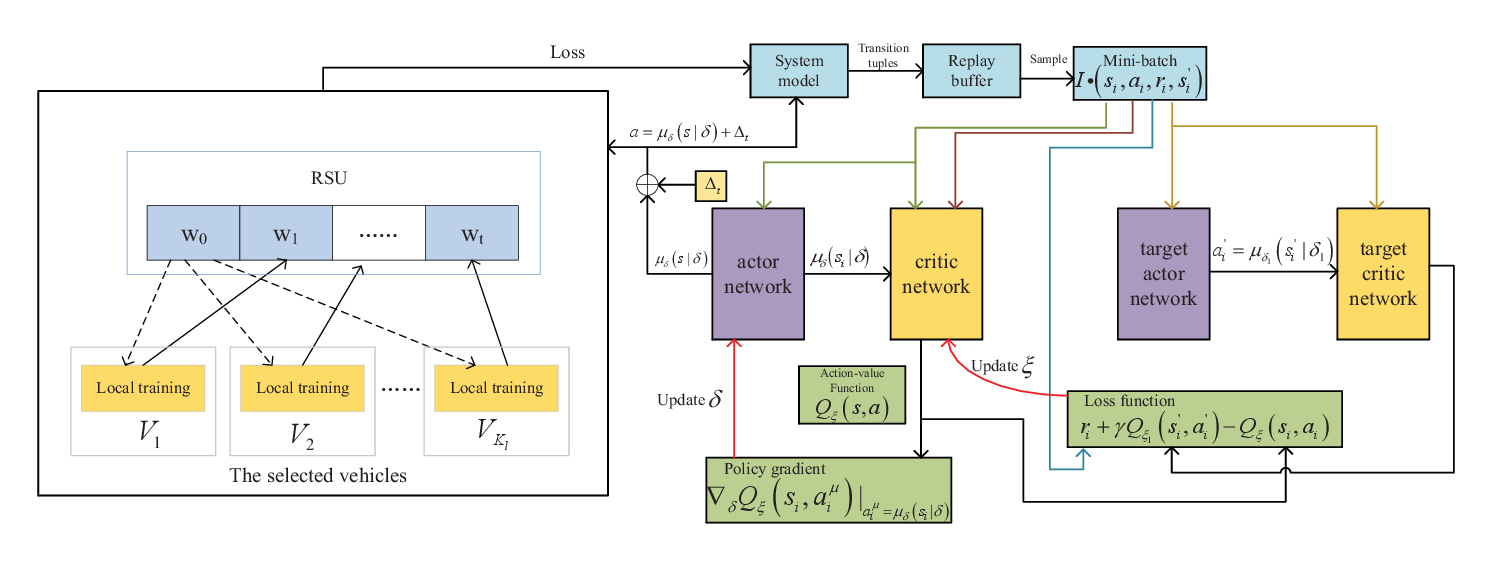}
	\caption{DDAFL flow diagram}
	\label{fig2}
\end{figure*}

The critic network employs gradient descent to update parameter $\xi $ for the purpose of minimizing $L\left( \xi  \right)$. Similarly, the actor network employs gradient ascent to update parameter $\delta $ for the purpose of maximizing $J\left( {{\mu _\delta }} \right)$.

Here the RSU calculates the gradient above, i.e., ${\nabla _\delta }J\left( {{\mu _\delta }} \right)$ based on $Q_{\xi}$ which is critic network's approximation result :
\begin{equation}
	\begin{split}
		% &\quad
		&\nabla_{\delta}J(\mu_{\delta})\\
		&\approx \frac{1}{I}\sum_{i=1}^{I}\nabla_{\delta}Q_{\xi}(s_i,a_{i}^{\mu})|_{a_{i}^{\mu}=\mu_{\delta}(s_i|\delta)}\\
		&=\frac{1}{I}\sum_{i=1}^{I}\nabla_{a_{i}^{\mu}}Q_{\xi}(s_i,a_{i}^{\mu})|_{a_{i}^{\mu}=\mu_{\delta}(s_i|\delta)} \\&\quad\cdot \nabla_{\delta}\mu_{\delta}(s_i|\delta)
	\end{split},
	\label{eq17}
\end{equation}
where we use chain rule to derive  ${\nabla _\delta }J\left( {{\mu _\delta }} \right)$ due to that $a_i^\mu$ represents ${Q_{\xi}(s_i,a_{i}^{\mu})}$'s input.

The target networks' parameters are updated as
\begin{equation}
	{\xi _1} \leftarrow \tau \xi  + \left( {1 - \tau } \right){\xi _1},
	\label{eq18}
\end{equation}
\begin{equation}
	{\delta _1} \leftarrow \tau \delta  + \left( {1 - \tau } \right){\delta _1},
	\label{eq19}
\end{equation}
the $\tau $ above represents a fixed value and $\tau  \ll 1$.

At last, the RSU puts $s'$ to the actor network and starts the next round.
One episode will be completed once time slot $t$ reaches $N$.
After that, the RSU initializes $s\left( t \right) = \left( {R\left( t \right),\mu \left( t \right),{d_x}\left( t \right),a\left( t-1 \right)} \right)$ again, and repeat the above procedures for the next episode.
Once episode's number reaches ${E_{\max }}$, the training will be finished and the RSU can obtain optimization parameters including ${\delta ^ {'} }$, ${\xi ^ {'} }$, $\delta _1^ {'} $, $\xi _1^ {'} $.
The process of our scheme named DDAFL is shown in Figure \ref{fig2}.

%\begin{figure*}
%\centering
%%\includegraphics[scale=0.55]{DAFL.eps}
%\includegraphics[width=16cm]{DAFL.eps}
%\caption{DDPG flow diagram}
%\label{fig2}
%\end{figure*}
\subsection{Testing stage}
During the testing stage, networks including critic, target actor, and target critic used during training stage are omitted. Instead, the optimized parameters ${\delta ^ {'}}$ are employed to implement the optimal policy.Therefore, neural networks can select an optimal strategy based on the trained parameters to select vehicles suitable for AFL, excluding those that have suffered Byzantine attacks and are not suitable for training under their own conditions.

 After the testing stage, we have selected vehicles to decrease the training delay, transmission delay and the loss of AFL.

\subsection{Enhanced Vehicle selection against Byzantine attack}
 In our model, Byzantine attacks occur on certain vehicles, and there are also two types of Byzantine attacks. When attacking a user's vehicle, both the user's local data and data labels may be tampered with. The former is called data flip in our simulation, while the latter is called class flip. Once a vehicle is attacked, they will incur significant losses during local AFL training. Moreover, Byzantine attacks will affect the accuracy of the global model after uploading to the global model, which in turn will affect the accuracy of all vehicles covered by the global model. In our model, threshold filtering is used to exclude vehicles that have been attacked, and in this part of the parameters$\beta _R$ is used to define the scope of filtering, which means that once the loss of the local model is much larger than that of the unaffected model, the model can be considered to have been attacked. Therefore, the corresponding ${a_{dn}}\left( t \right)$ is set as 0 and the vehicle will be not selected.
Thus, we propose a threshold screening method to further select vehicles against Byzantine attack.

Specifically, the RSU has a trusted dataset ${D_{RSU}}$, i.e., the dataset that does not be affected by Byzantine attack.
The RSU trains its local model by using ${D_{RSU}}$ while all vehicle train their own model and calculates the loss of its local model ${L_{RSU}}$.
When a vehicle uploads its local model, if its model loss ${L_{{w_k}}}$ is lower than that of RSU ${L_{RSU}}$, i.e.,
${L_{{w_k}}} \le {\beta _R} \cdot {L_{RSU}}$ (where ${\beta _R}$ is a parameter), the vehicle can take part in the AFL.
Otherwise, it will be dropped.
The specific algorithm is demonstrated in Algorithm \ref{al4} and we take vehicle ${V_{Dk}}$ as an example.
\begin{algorithm}[h]
	\small
	\caption{Byzantine fault-tolerant filtering AFL algorithm}
	\label{al4}
	Set up global model $w_0$;\\
	\For{round $x$ from $1$ to ${E_{pi}}$}
	{
		$w_k \leftarrow \textbf{Vehicle Updates}(w)$;\\
		Compute ${L_{{w_k}}}$ according to Eq. \eqref{eq20};\\
		Get local model ${w_{Dkw}}$ of vehicle ${V_{Dk}}$ according to Eq. \eqref{eq25};\\
		${V_{Dk}}$ uploads ${L_{{w_k}}}$ and ${w_{Dkw}}$ to RSU;\\
		RSU compute local model ${w_{RSU}}$ according to Eq. \eqref{eq21};\\
		RSU compute loss ${L_{RSU}}$ of ${w_{RSU}}$ according to Eq. \eqref{eq20};\\
		\If {${L_{{w_k}}} \le {\beta _R} \cdot {L_{RSU}}$}{global model get updated according to Eq. \eqref{eq26};}
	}
	
	\textbf{Vehicle Update}($w$):\\
	\textbf{Input:} the newest global model that downloaded from RSU;\\
	\For{local training process $m$ from $1$ to $L$}
	{
		${V_{Dk}}$ computes cross-entropy loss according to Eq. \eqref{eq20};\\
		${V_{Dk}}$ updates the local model according to Eq. \eqref{eq21};\\
	}
	\Return Updated local model of ${V_{Dk}}$
	
\end{algorithm}

\section{Simulation Results}
In this section, we will assess the effectiveness of the scheme we provided through simulation.

\subsection{Simulation setup}
Python 3.9 is used in our paper. In the DDPG, both the actor network and critic network employ DNNs which has two hidden layers. The first contains 400 neurons, while the second contains 300 neurons. Additionally, the exploration noise follows Ornstein-Uhlenbeck with 0.02 variance and 0.15 decay rate.

In the simulation,we use the Mnist dataset as the data source for vehicles and RSUs, and ensure that the clean dataset for RSU is consistent in size and type with the dataset for normal vehicles by setting an additional vehicle as the clean dataset source for RSU. The available computational capabilities follows the truncated normal distribution. The unit of available computational capabilities refers to cycles per second.The channel gain ${h_n}\left( t \right)$ is predicted based on the autoregressive model, which is in line with most vehicle networking scenarios.
In the part of vehicle selection based on DDPG, we set one bad vehicle. Specifically, it has less information, lower computational capabilities. In addition, the local model provided will be affected by noise.
Since we only considered the Byzantine attack for already selected vehicles, the Byzantine attack has not been considered in the vehicle selection based on DDPG.
After vehicles are selected based on DDPG algorithm, two of the selected vehicles will be affected by Byzantine attack.
We set up two different classes of Byzantine attacks.
The class flip will attack the label of data on the vehicles which label ${y_a}$ will be replaced by $9 - {y_a}$.
The data flip will attack the data on the vehicles which data $a$ will be replaced by $1-a$.

We compared our scheme with traditional FL and AFL, and used ablation experiments to demonstrate the importance of certain factors we considered, such as upload delay. We also removed the consideration of Byzantine attacks in our scheme to demonstrate the significance of defending against Byzantine attacks. Here we use test error rate denoting the rate of clean dataset being misclassified. Some other simulation parameters are shown in Table \ref{tab}.

\begin{table}\footnotesize
	\caption{The Parameter used in simulation}
	\label{tab}
	\centering
	\begin{tabular}{|c|c|c|c|}
		%\toprule
		\hline
		\textbf{Parameters} &\textbf{Value} &\textbf{Parameters} &\textbf{Value}\\
		\hline
		$v$ & 20 & $\Lambda$ & 7$m$ \\
		\hline
		$t$ & 0.5 & $K$ & 5 \\
		\hline
		$I$ & 64 & $p_0$ & 0.25$w$ \\
		\hline
		$E_{m}$ & 1000 & ${{\sigma }^{2}}$ & ${{10}^{-9}}mw$ \\
		\hline
		$E_{m }^{'}$ & 3 & $H_R$ & 10$m$ \\
		\hline
		$B$ & 1000$HZ$ & $\left| w \right|$ & 5000$bits$ \\
		\hline
		$\gamma$ & 0.99$m/s$ & $\alpha$ & 2 \\
		\hline
		$C_0$ & ${{10}^{6}}CPU-cycles$ & $M_1$ & 0.9 \\
		\hline
		$\tau$ & 0.001 & $M_2$ & 0.9 \\
		\hline
		$d_y$ & 5$m$ & ${\beta _R}$ & 1.25 \\
		\hline
	\end{tabular}
\end{table}
\subsection{Results}
\begin{figure}[h]
	\center
	\includegraphics[scale=0.55]{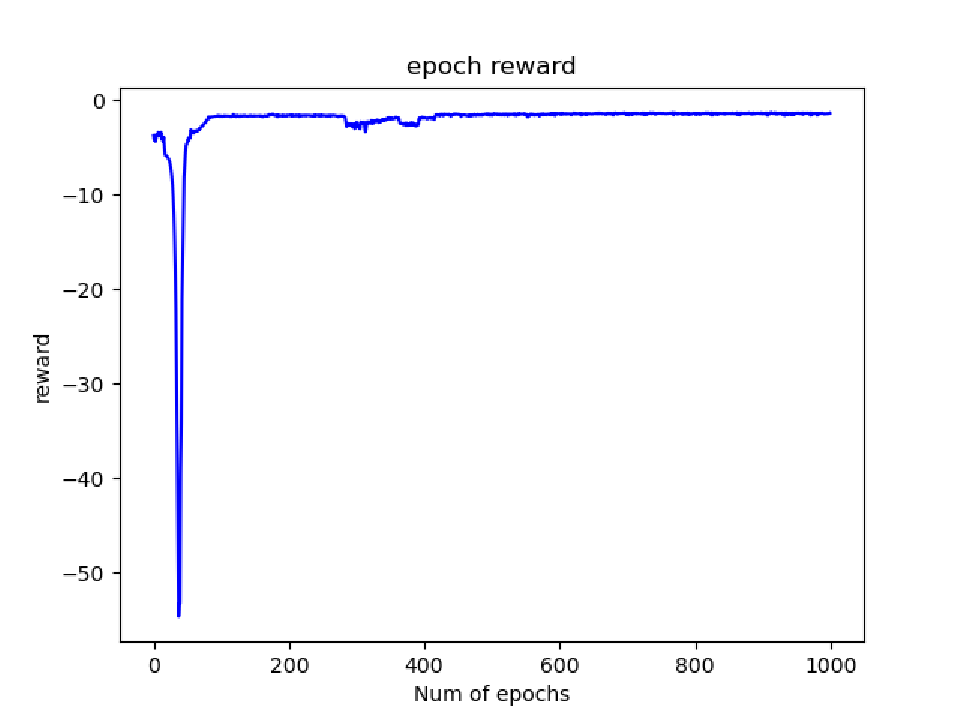}
	\caption{Epoch reward in training stage}
	\label{fig3}
\end{figure}

From Fig. \ref{fig3} we can see that with the increase of number of epochs, the reward will fluctuate and then gradually become stable. This means that the system has acquired the most effective strategy. In addition, the neural network's training process is finished.
\begin{figure}[ht]
	\center
	\includegraphics[scale=0.4]{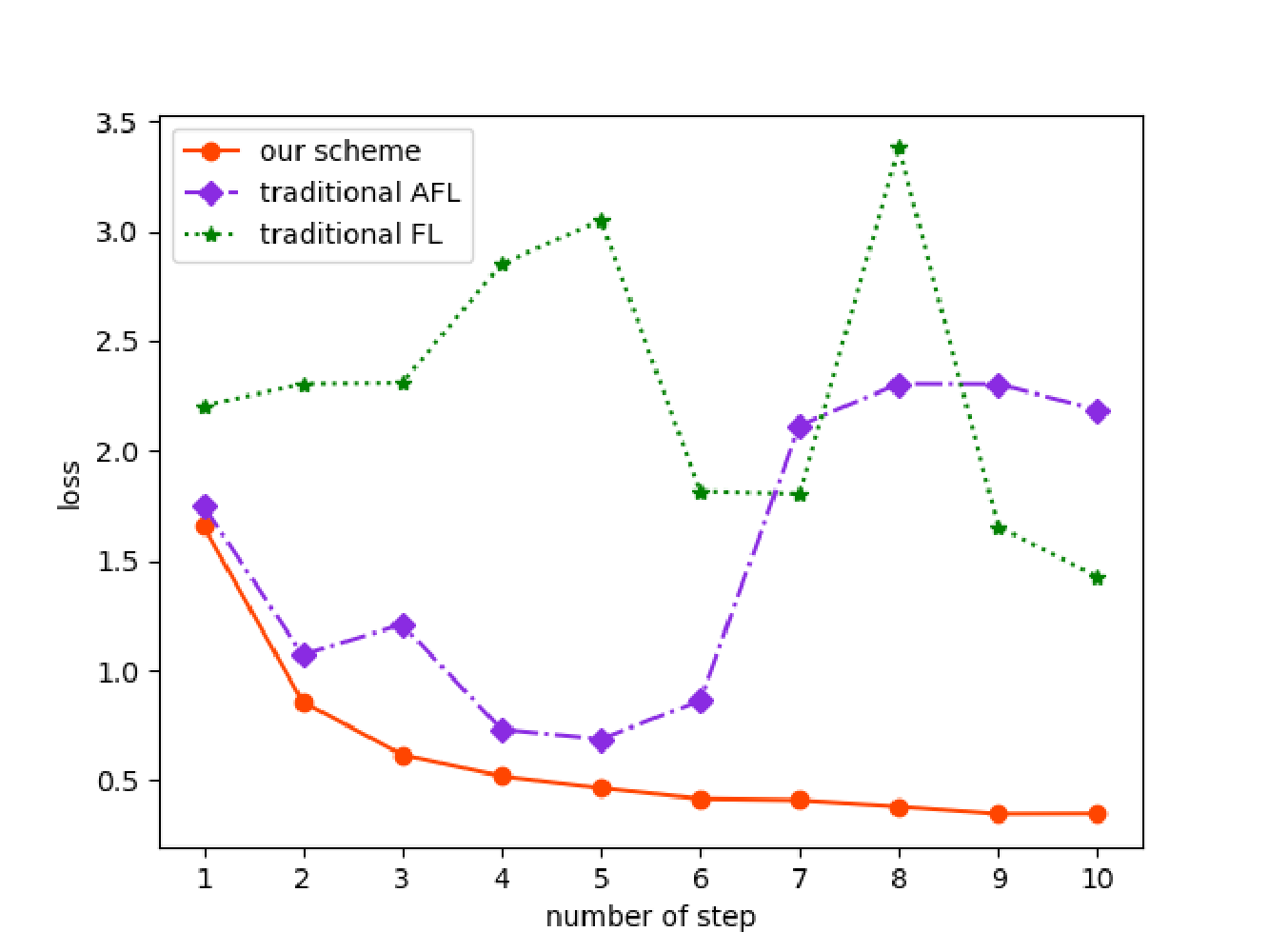}
	\caption{Loss in testing stage with bad node}
	\label{fig4}
\end{figure}
\begin{figure}
	\center
    
	\includegraphics[scale=0.4]{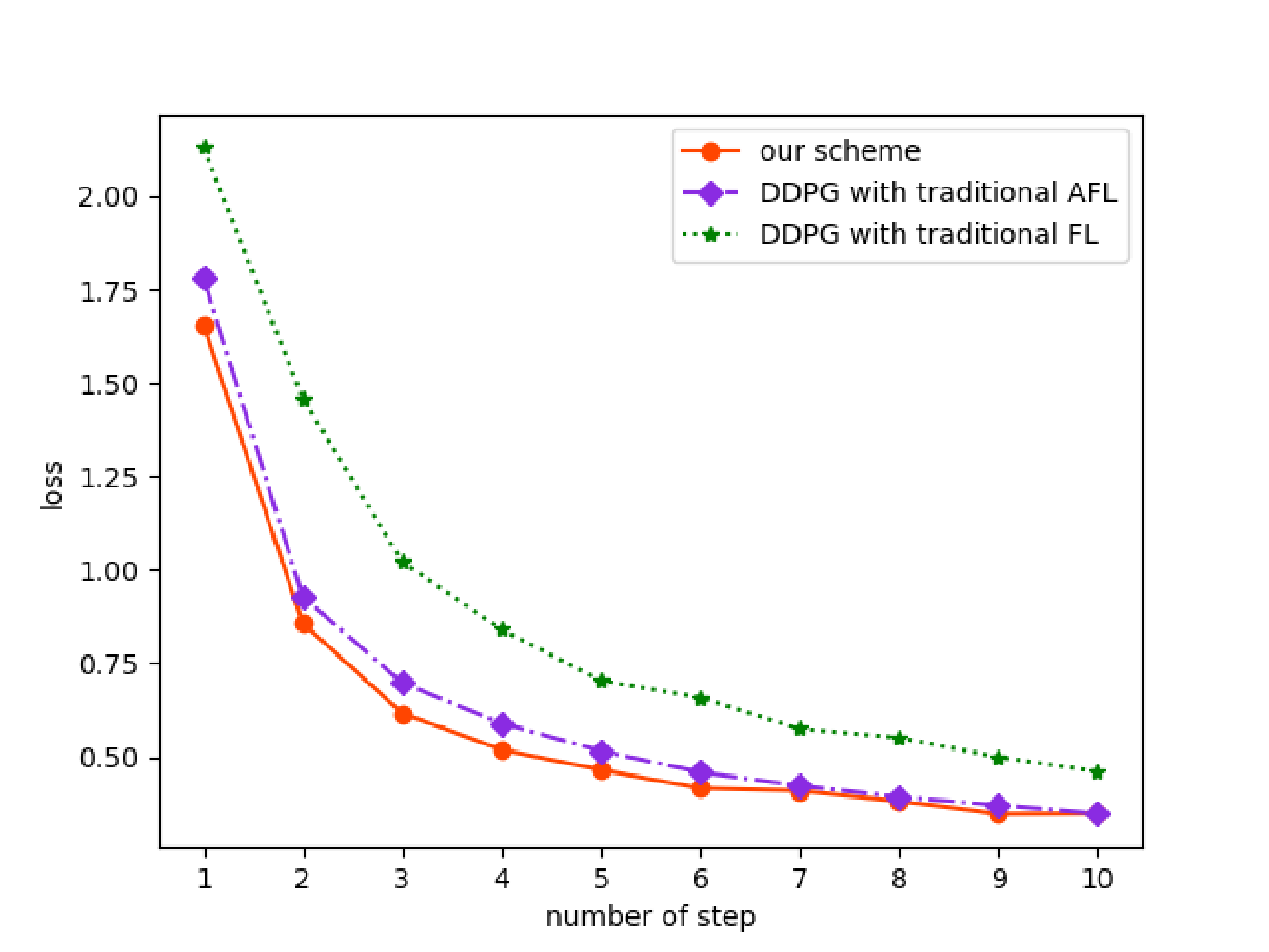}
	\caption{Loss in testing stage without bad node}
	\label{fig5}
\end{figure}
Fig. \ref{fig4} investigates the loss of different methods, including the method we provided, conventional AFL and FL during the testing stage with the existence of bad node.
It is shown that the loss of our scheme keeps decreasing and finally becomes stable.This is because using the DDPG algorithm to select vehicles will avoid selecting those bad nodes to participate in AFL. On the contrary, the two federated learning algorithms that do not use DDPG will inevitably choose bad nodes when selecting vehicles. This is the reason why the loss of traditional AFL and traditional FL fluctuates sharply, which indicates that we have effectively screened out bad nodes.

Fig. \ref{fig5} shows the comparison of the loss of different schemes in the testing stage without bad node.
From the figure it can be seen that our proposed scheme has the lowest loss and outperforms two other baseline algorithms. This is because we have considered the local training and transmission latency’s influence.In addition, since our scheme considers weighted aggregation during global aggregation, some nodes with relatively poor performance will have a smaller impact on the global model during aggregation, resulting in smaller loss, which traditional AFL and FL do not have.
\begin{figure}[h]
	\center
	\includegraphics[scale=0.4]{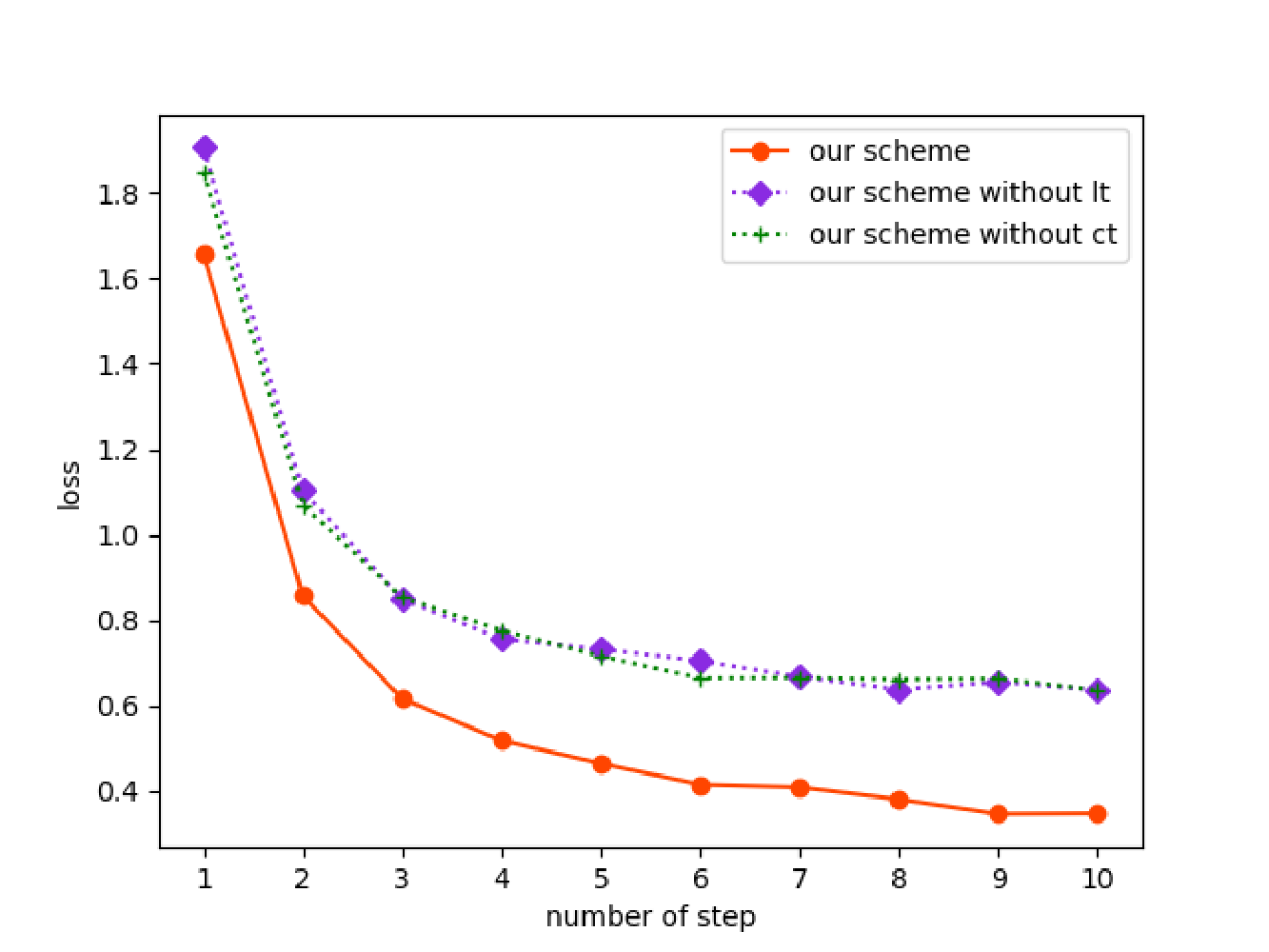}
	\caption{Loss in testing stage comparison}
	\label{fig6}
\end{figure}
\begin{figure}[ht]
	\center
	\includegraphics[scale=0.4]{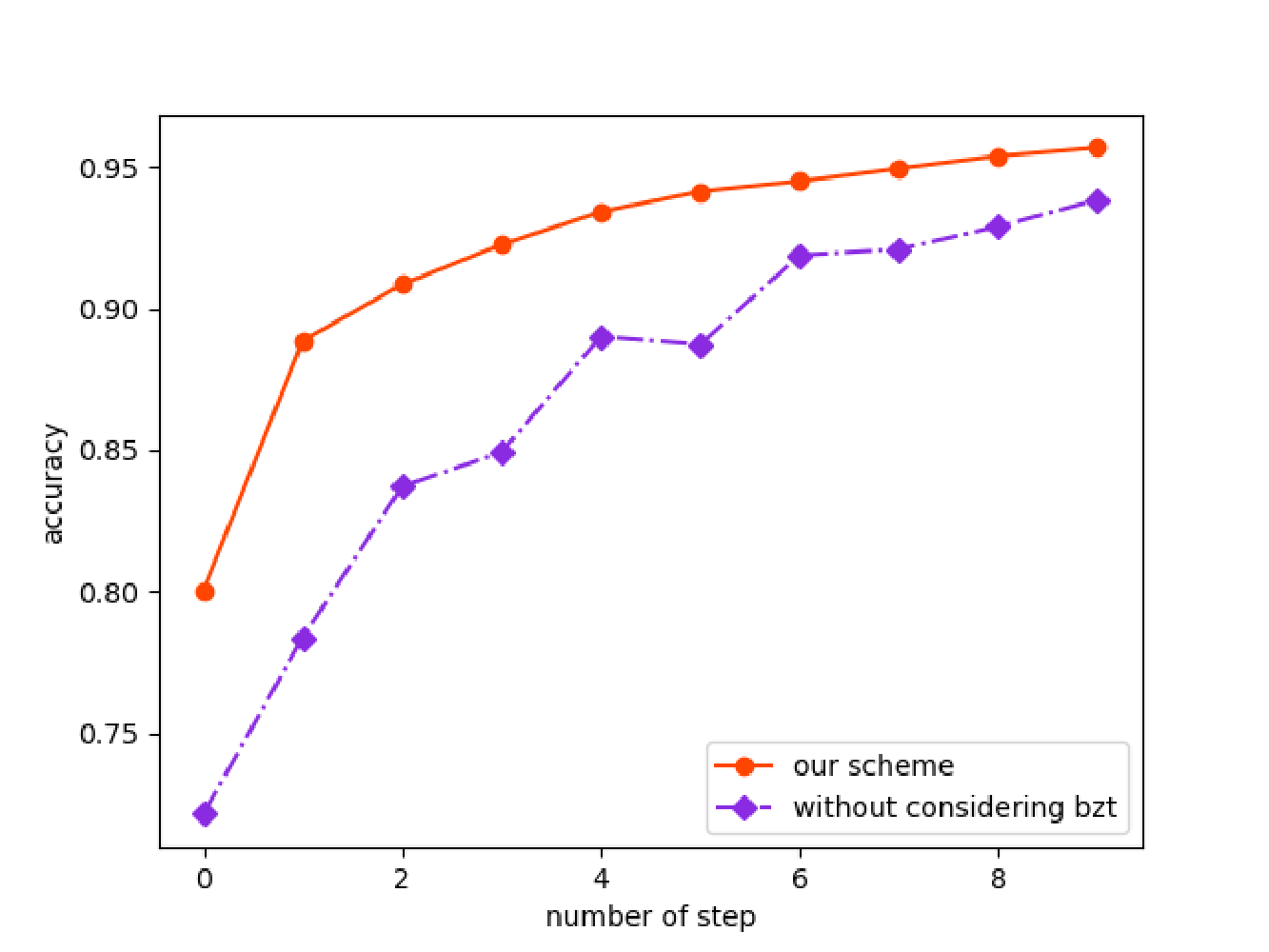}
	\caption{Accuracy under class flip}
	\label{fig7}
\end{figure}

Fig.6 used ablation experiments to eliminate the effects of local computing time and transmission delay in our proposed scheme, and then compared it with our complete scheme to demonstrate the effects of factors such as vehicle mobility and computing resources In the figure,$lt$ represents local training time, $ct$ represents communication time. The local training time is affected by the available computing resources and data volume of the vehicle, while the communication time is also the model upload delay, which is affected by channel conditions and vehicle mobility. 
According to the figure, we can find that ignoring either of these two will lead to an increase in the overall loss of the scheme. This is because a long local training time can cause the vehicle model to upload too slowly, resulting in outdated information.

\begin{figure}[t]
	\center
	\includegraphics[scale=0.4]{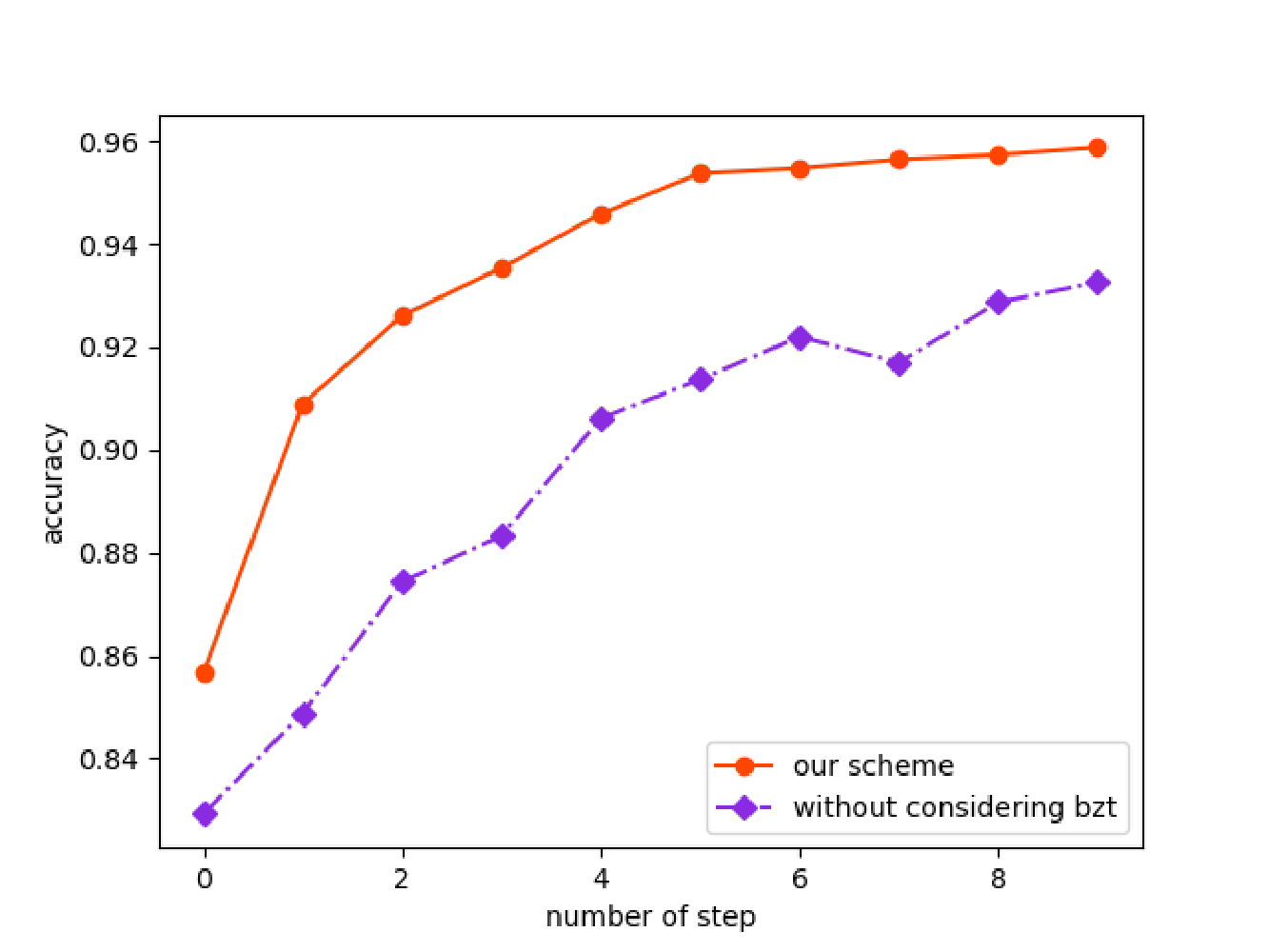}
	\caption{Accuracy under data flip}
	\label{fig8}
\end{figure}
\begin{figure}[h]
	\center
	\includegraphics[scale=0.4]{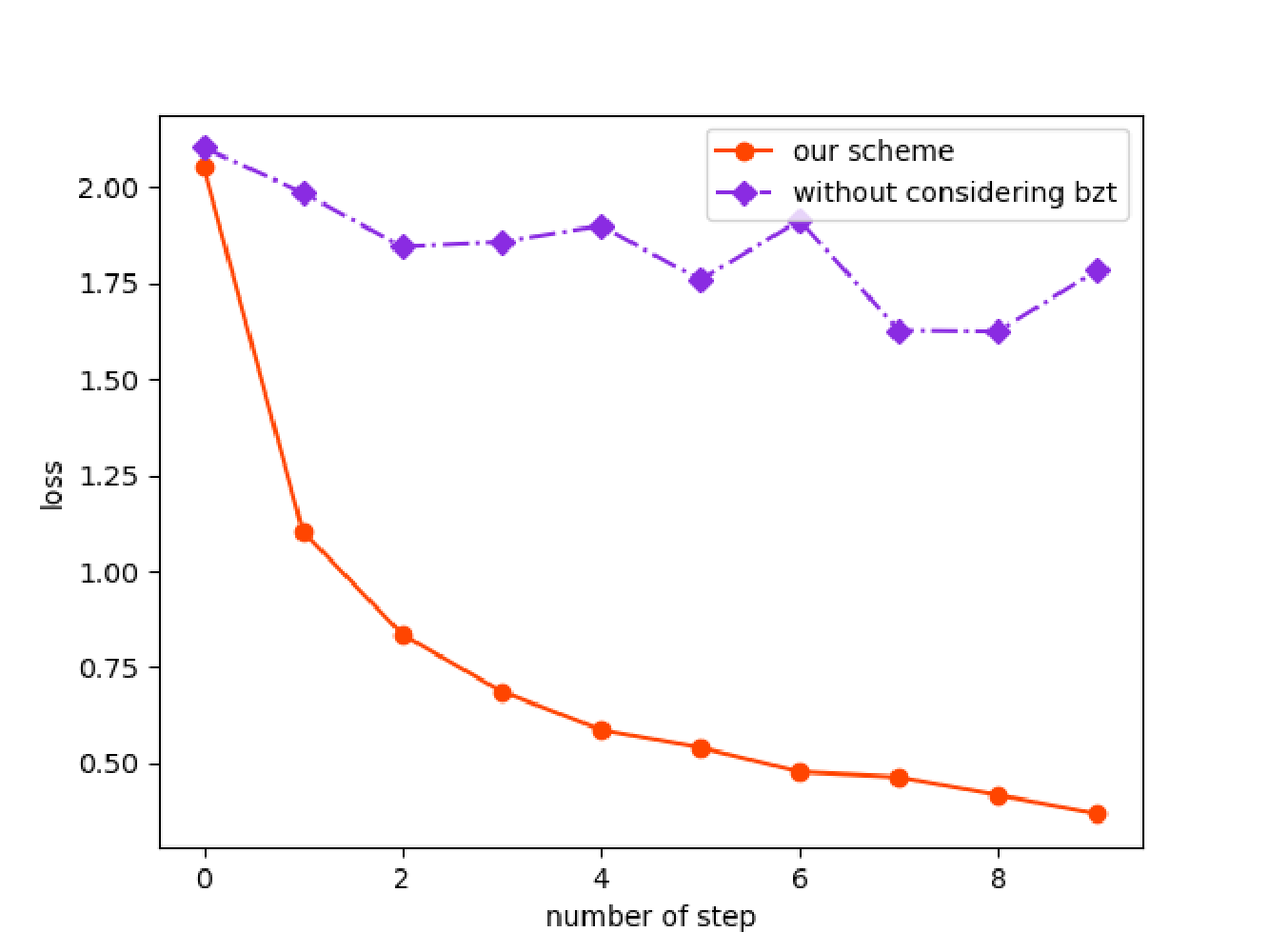}
	\caption{Loss under class flip}
	\label{fig9}
\end{figure}
Similarly, a long upload time can also cause the vehicle model information to become outdated, affecting the accuracy of the global model. Therefore, from the loss curve, the performance difference between the two schemes is not significant. Therefore, we can draw that considering factors such as vehicle mobility, computing power, and channel conditions is important.
\begin{figure}
	\center
	\includegraphics[scale=0.4]{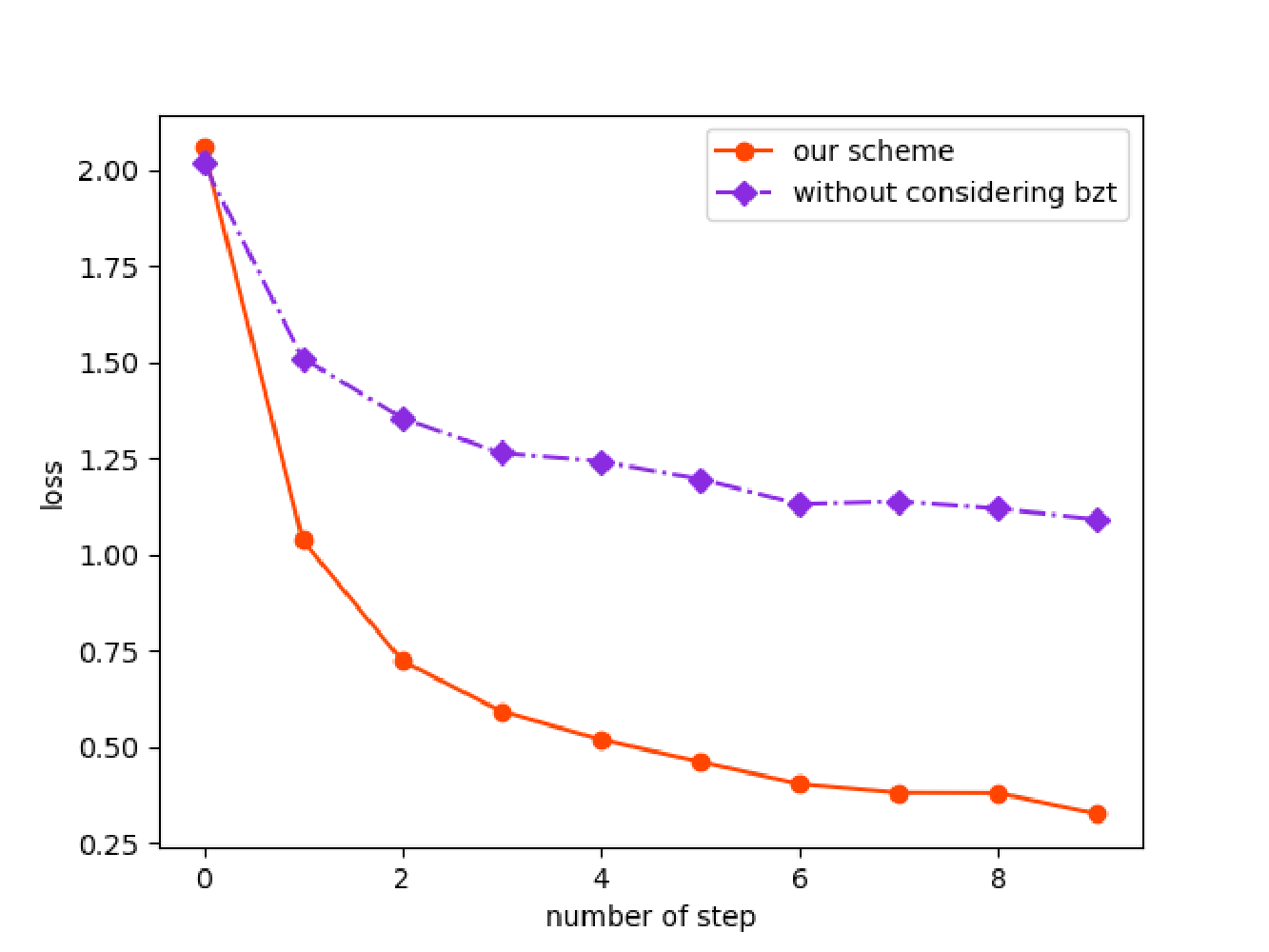}
	\caption{Loss under data flip}
	\label{fig10}
\end{figure}
\begin{figure}[h]
	\center
	\includegraphics[scale=0.4]{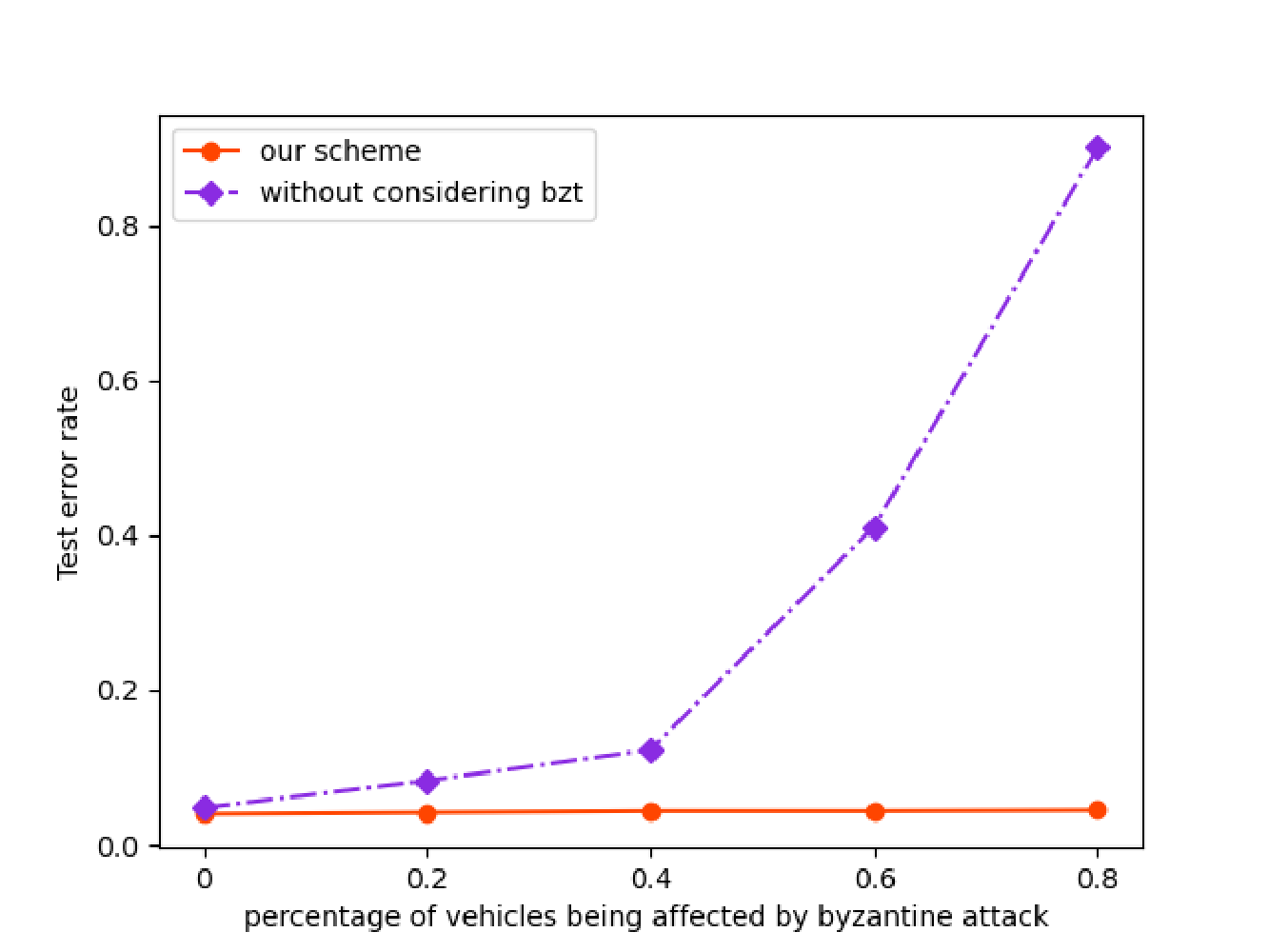}
	\caption{Test error rate under class flip}
	\label{fig11}
\end{figure}

Fig. \ref{fig7} and Fig. \ref{fig8} show the accuracy of our scheme and the scheme without considering Byzantinge attacks under different attacks, i.e., class flip and data flip.
As shown in the chart, we can know that both methods' precision increase as steps' number raises. This is because when more and more local models are received at RSU in different steps, the number of global model updates keeps increasing, improving global model's precision.
Simultaneously, it can be seen that our scheme has a higher accuracy and is more stable while the scheme without considering Byzantine attacks has a lower accuracy. This is because our scheme takes into account the factor of Byzantine attacks. As our complete scheme filters out vehicles that have been attacked by Byzantine attacks before AFL, the accuracy of the global model will not be affected by Byzantine attacks. However, once our scheme eliminates the threshold filtering process, it will cause the local model that has been attacked to be updated by the global model, which affects the accuracy of the global model. Numerically, although our model accuracy is high, it is not very obvious. This is because we did not set a large number of attacked vehicles in this experiment, so the impact is not significant. The impact on the number of attacked vehicles will be shown in the following experiments. In addition, it can be seen that different attacks have different impacts on global model’s precision. The accuracy of models subjected to class flip attacks is generally lower than those subjected to data attacks, as the impact of class flip in the dataset is greater than that of data flip.

\begin{figure}[t]
	\center
	\includegraphics[scale=0.4]{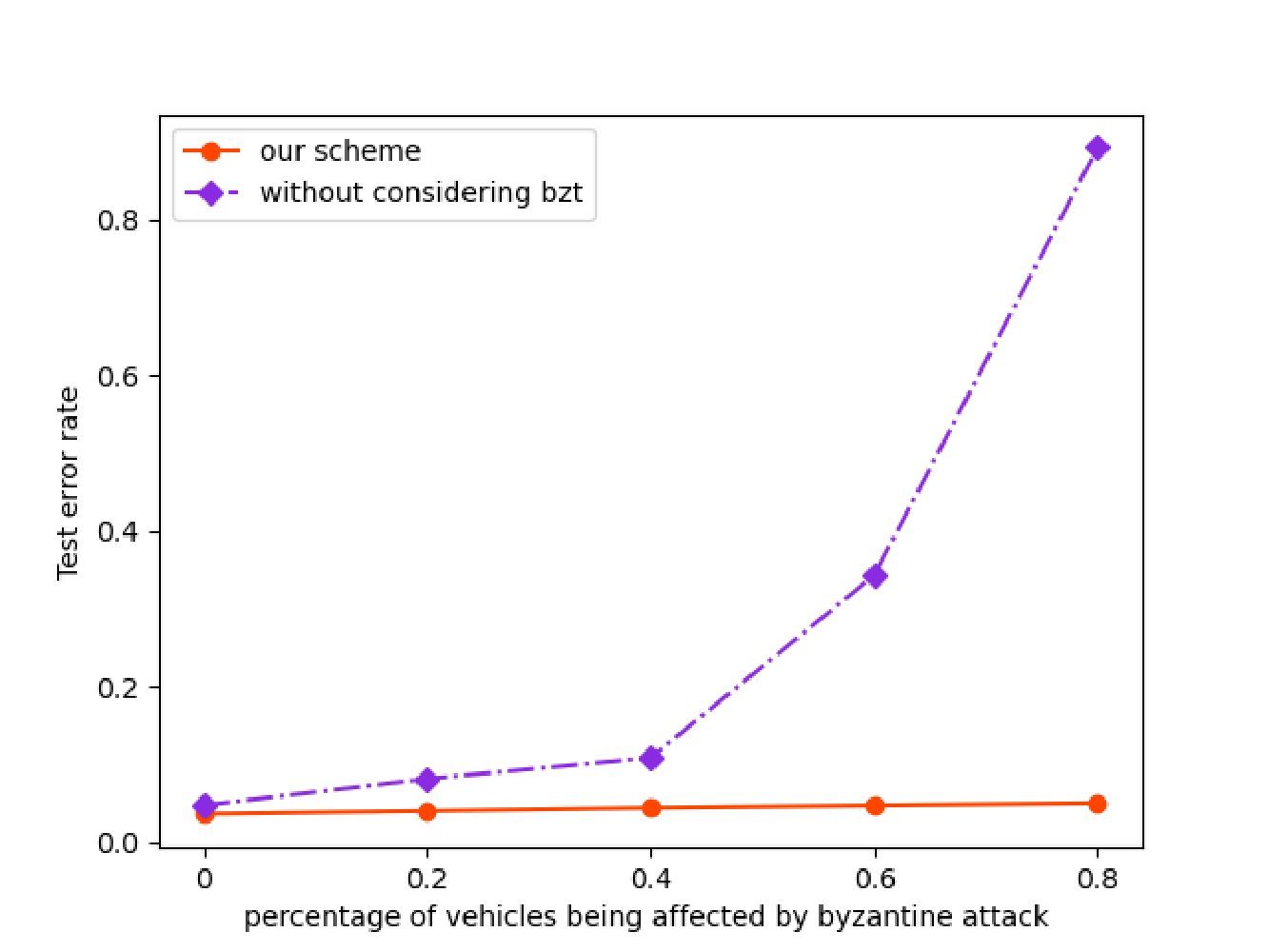}
	\caption{Test error rate under data flip}
	\label{fig12}
\end{figure}
Fig. \ref{fig9} and Fig. \ref{fig10} show the loss of our scheme and the scheme without considering Byzantine attacks under different attacks. From the figure it can be seen that the loss of both schemes decrease as the number of steps increases. However, the consequences of class reversal attacks are more severe, with significant fluctuations in losses and relatively small decreases during iterations. This is because class reversal destroys the labels of the dataset, leading to more severe confusion in the dataset. Meanwhile, our approach exhibits a less loss compared to the method without considering Byzantine attacks, due to the fact that our plan takes into account Byzantine attacks and eliminates the involvement of attacked vehicles in AFL in advance, which results in lower losses and a steady decrease.

Fig. \ref{fig11} and Fig. \ref{fig12}show the test error rate of our scheme and the scheme without considering Byzantine attacks under different attacks. It can be seen that the test error rate of our scheme is always lower than that of the scheme without considering Byzantine attacks with the increase of percentage of vehicles being affected by Byzantine attack. Meanwhile, as the percentage of vehicles affected by Byzantine attacks increases, the test error rate of the scheme without considering Byzantine attacks will rapidly increase. This is because our scheme identify and exclude vehicles affected by Byzantine attacks in advance, so that they would not affect the global model. Therefore, increasing the number of vehicles affected by Byzantine attacks almost does not affect the accuracy and error rate of the global model. For the scheme that has not considered Byzantine attacks, the error rate of the global model is acceptable when the percentage of attacked vehicles is small. However, once the percentage of attacked vehicles is large, the error rate will rapidly increase. This is because when the percentage is small, most of the data is normal, and the degree of model impact is small. However, once the percentage of attacked vehicles is large, it will cause the parameters of the model to be rapidly affected, resulting in a rapid increase in the error rate.

\section{Conclusion}
In this paper, the vehicle mobility, computational resources, amount of local data, external interference and Byzantine attack have been taken into account to optimize the vehicle selection scheme in AFL. The findings can be concluded in the following manner:

\begin{itemize}
	\item In terms of precision,the scheme we provided is more accurate compared to the traditional AFL and FL, due to the fact that the bad nodes are removed efficiently in our system, so that the bad nodes can not affect the updating process of global model.

	\item Although the bad nodes have been eliminated, in terms of precision, the method we provided performs still better compared to traditional AFL and FL, the reason is that  the  vehicles' mobility, computational capability, amount of local data are taken into account to optimize local model's weight  during the process of global model's aggregation, which improves global model's precision.
	\item Our scheme outperforms the scheme without considering Byzantine attack which is attributed to the fact that the proposed scheme has selected the vehicles to prevent the Byzantine attack from impacting the test error rate.
\end{itemize}

\bibliographystyle{IEEEtran}
\bibliography{ref}

\begin{thebibliography}{10}
\providecommand{\url}[1]{#1}
\csname url@samestyle\endcsname
\providecommand{\newblock}{\relax}
\providecommand{\bibinfo}[2]{#2}
\providecommand{\BIBentrySTDinterwordspacing}{\spaceskip=0pt\relax}
\providecommand{\BIBentryALTinterwordstretchfactor}{4}
\providecommand{\BIBentryALTinterwordspacing}{\spaceskip=\fontdimen2\font plus
\BIBentryALTinterwordstretchfactor\fontdimen3\font minus
  \fontdimen4\font\relax}
\providecommand{\BIBforeignlanguage}[2]{{%
\expandafter\ifx\csname l@#1\endcsname\relax
\typeout{** WARNING: IEEEtran.bst: No hyphenation pattern has been}%
\typeout{** loaded for the language `#1'. Using the pattern for}%
\typeout{** the default language instead.}%
\else
\language=\csname l@#1\endcsname
\fi
#2}}
\providecommand{\BIBdecl}{\relax}
\BIBdecl

\bibitem{1}
W.~Wang, F.~Xia, H.~Nie, Z.~Chen, Z.~Gong, X.~Kong, and W.~Wei, ``Vehicle
  trajectory clustering based on dynamic representation learning of internet of
  vehicles,'' \emph{IEEE Transactions on Intelligent Transportation Systems},
  vol.~22, no.~6, pp. 3567--3576, 2020.

\bibitem{2}
B.~Yin, Y.~Wu, T.~Hu, J.~Dong, and Z.~Jiang, ``An efficient collaboration and
  incentive mechanism for internet of vehicles (iov) with secured information
  exchange based on blockchains,'' \emph{IEEE Internet of Things Journal},
  vol.~7, no.~3, pp. 1582--1593, 2019.

\bibitem{46}
J.~Zhao, Q.~Li, X.~Ma, and F.~R. Yu, ``Computation offloading for edge
  intelligence in two-tier heterogeneous networks,'' \emph{IEEE Transactions on
  Network Science and Engineering}, 2023.

\bibitem{3}
A.~Cardaillac and M.~Ludvigsen, ``A communication interface for multilayer
  cloud computing architecture for low cost underwater vehicles,''
  \emph{IFAC-PapersOnLine}, vol.~55, no.~14, pp. 77--82, 2022.

\bibitem{50}
S.~Zhang and N.~Ansari, ``Latency aware 3d placement and user association in
  drone-assisted heterogeneous networks with fso-based backhaul,'' \emph{IEEE
  Transactions on Vehicular Technology}, vol.~70, no.~11, pp. 11\,991--12\,000,
  2021.

\bibitem{4}
J.~Lee and W.~Na, ``A survey on vehicular edge computing architectures,'' in
  \emph{2022 13th International Conference on Information and Communication
  Technology Convergence (ICTC)}.\hskip 1em plus 0.5em minus 0.4em\relax IEEE,
  2022, pp. 2198--2200.

\bibitem{5}
Y.-J. Ku, P.-H. Chiang, and S.~Dey, ``Real-time qos optimization for vehicular
  edge computing with off-grid roadside units,'' \emph{IEEE Transactions on
  Vehicular Technology}, vol.~69, no.~10, pp. 11\,975--11\,991, 2020.

\bibitem{52}
Q.~Wu, S.~Wang, H.~Ge, P.~Fan, Q.~Fan, and K.~B. Letaief, ``Delay-sensitive
  task offloading in vehicular fog computing-assisted platoons,'' \emph{IEEE
  Transactions on Network and Service Management}, 2023.

\bibitem{7}
X.~Yuan, J.~Chen, N.~Zhang, X.~Fang, and D.~Liu, ``A federated bidirectional
  connection broad learning scheme for secure data sharing in internet of
  vehicles,'' \emph{China Communications}, vol.~18, no.~7, pp. 117--133, 2021.

\bibitem{8}
B.~Gu, A.~Xu, Z.~Huo, C.~Deng, and H.~Huang, ``Privacy-preserving asynchronous
  vertical federated learning algorithms for multiparty collaborative
  learning,'' \emph{IEEE transactions on neural networks and learning systems},
  vol.~33, no.~11, pp. 6103--6115, 2021.

\bibitem{9}
Q.~Wu, Y.~Zhao, Q.~Fan, P.~Fan, J.~Wang, and C.~Zhang, ``Mobility-aware
  cooperative caching in vehicular edge computing based on asynchronous
  federated and deep reinforcement learning,'' \emph{IEEE Journal of Selected
  Topics in Signal Processing}, vol.~17, no.~1, pp. 66--81, 2022.

\bibitem{10}
Q.~Wu, S.~Xia, P.~Fan, Q.~Fan, and Z.~Li, ``Velocity-adaptive v2i fair-access
  scheme based on ieee 802.11 dcf for platooning vehicles,'' \emph{Sensors},
  vol.~18, no.~12, p. 4198, 2018.

\bibitem{51}
Q.~Wu, W.~Wang, P.~Fan, Q.~Fan, J.~Wang, and K.~B. Letaief, ``Urllc-awared
  resource allocation for heterogeneous vehicular edge computing,'' \emph{IEEE
  Transactions on Vehicular Technology}, 2024.

\bibitem{11}
W.~Qiong, S.~Shuai, W.~Ziyang, F.~Qiang, F.~Pingyi, and Z.~Cui, ``Towards v2i
  age-aware fairness access: a dqn based intelligent vehicular node training
  and test method,'' \emph{Chinese Journal of Electronics}, vol.~32, no.~6, pp.
  1230--1244, 2023.

\bibitem{47}
J.~Zhao, X.~Xiong, Q.~Zhang, and D.~Wang, ``Extended multi-component gated
  recurrent graph convolutional network for traffic flow prediction,''
  \emph{IEEE Transactions on Intelligent Transportation Systems}, 2023.

\bibitem{12}
X.~Chen, W.~Wei, Q.~Yan, N.~Yang, and J.~Huang, ``Time-delay deep q-network
  based retarder torque tracking control framework for heavy-duty vehicles,''
  \emph{IEEE Transactions on Vehicular Technology}, vol.~72, no.~1, pp.
  149--161, 2022.

\bibitem{13}
Y.~M. Saputra, D.~N. Nguyen, D.~T. Hoang, and E.~Dutkiewicz, ``Selective
  federated learning for on-road services in internet-of-vehicles,'' in
  \emph{2021 IEEE Global Communications Conference (GLOBECOM)}.\hskip 1em plus
  0.5em minus 0.4em\relax IEEE, 2021, pp. 1--6.

\bibitem{14}
N.~Plewtong and B.~DeBruhl, ``Game theoretic analysis of a byzantine attacker
  in vehicular mix-zones,'' in \emph{Decision and Game Theory for Security: 9th
  International Conference, GameSec 2018, Seattle, WA, USA, October 29--31,
  2018, Proceedings 9}.\hskip 1em plus 0.5em minus 0.4em\relax Springer, 2018,
  pp. 277--295.

\bibitem{15}
Q.~Wu, X.~Wang, Q.~Fan, P.~Fan, C.~Zhang, and Z.~Li, ``High stable and accurate
  vehicle selection scheme based on federated edge learning in vehicular
  networks,'' \emph{China Communications}, vol.~20, no.~3, pp. 1--17, 2023.

\bibitem{16}
Y.~Chu, Z.~Wei, X.~Fang, S.~Chen, and Y.~Zhou, ``A multiagent federated
  reinforcement learning approach for plug-in electric vehicle fleet charging
  coordination in a residential community,'' \emph{IEEE Access}, vol.~10, pp.
  98\,535--98\,548, 2022.

\bibitem{17}
L.~Liu, Z.~Xi, K.~Zhu, R.~Wang, and E.~Hossain, ``Mobile charging station
  placements in internet of electric vehicles: A federated learning approach,''
  \emph{IEEE Transactions on Intelligent Transportation Systems}, vol.~23,
  no.~12, pp. 24\,561--24\,577, 2022.

\bibitem{18}
F.~Liang, Q.~Yang, R.~Liu, J.~Wang, K.~Sato, and J.~Guo, ``Semi-synchronous
  federated learning protocol with dynamic aggregation in internet of
  vehicles,'' \emph{IEEE Transactions on Vehicular Technology}, vol.~71, no.~5,
  pp. 4677--4691, 2022.

\bibitem{19}
X.~Kong, H.~Gao, G.~Shen, G.~Duan, and S.~K. Das, ``Fedvcp: A
  federated-learning-based cooperative positioning scheme for social internet
  of vehicles,'' \emph{IEEE Transactions on Computational Social Systems},
  vol.~9, no.~1, pp. 197--206, 2021.

\bibitem{20}
X.~Li, L.~Lu, W.~Ni, A.~Jamalipour, D.~Zhang, and H.~Du, ``Federated
  multi-agent deep reinforcement learning for resource allocation of
  vehicle-to-vehicle communications,'' \emph{IEEE Transactions on Vehicular
  Technology}, vol.~71, no.~8, pp. 8810--8824, 2022.

\bibitem{21}
C.~Li, Y.~Zhang, and Y.~Luo, ``A federated learning-based edge caching approach
  for mobile edge computing-enabled intelligent connected vehicles,''
  \emph{IEEE Transactions on Intelligent Transportation Systems}, vol.~24,
  no.~3, pp. 3360--3369, 2022.

\bibitem{22}
S.~R. Pokhrel and J.~Choi, ``Improving tcp performance over wifi for internet
  of vehicles: A federated learning approach,'' \emph{IEEE transactions on
  vehicular technology}, vol.~69, no.~6, pp. 6798--6802, 2020.

\bibitem{23}
X.~Zhou, W.~Liang, J.~She, Z.~Yan, I.~Kevin, and K.~Wang, ``Two-layer federated
  learning with heterogeneous model aggregation for 6g supported internet of
  vehicles,'' \emph{IEEE Transactions on Vehicular Technology}, vol.~70, no.~6,
  pp. 5308--5317, 2021.

\bibitem{24}
P.~Lv, L.~Xie, J.~Xu, X.~Wu, and T.~Li, ``Misbehavior detection in vehicular ad
  hoc networks based on privacy-preserving federated learning and blockchain,''
  \emph{IEEE Transactions on Network and Service Management}, vol.~19, no.~4,
  pp. 3936--3948, 2022.

\bibitem{25}
T.~Zeng, O.~Semiari, M.~Chen, W.~Saad, and M.~Bennis, ``Federated learning on
  the road autonomous controller design for connected and autonomous
  vehicles,'' \emph{IEEE Transactions on Wireless Communications}, vol.~21,
  no.~12, pp. 10\,407--10\,423, 2022.

\bibitem{26}
C.~Pan, Z.~Wang, H.~Liao, Z.~Zhou, X.~Wang, M.~Tariq, and S.~Al-Otaibi,
  ``Asynchronous federated deep reinforcement learning-based urllc-aware
  computation offloading in space-assisted vehicular networks,'' \emph{IEEE
  Transactions on Intelligent Transportation Systems}, 2022.

\bibitem{27}
K.~Bedda, Z.~M. Fadlullah, and M.~M. Fouda, ``Efficient wireless network
  slicing in 5g networks: An asynchronous federated learning approach,'' in
  \emph{2022 IEEE International Conference on Internet of Things and
  Intelligence Systems (IoTaIS)}.\hskip 1em plus 0.5em minus 0.4em\relax IEEE,
  2022, pp. 285--289.

\bibitem{28}
Z.~Liu, C.~Guo, D.~Liu, and X.~Yin, ``An asynchronous federated learning
  arbitration model for low-rate ddos attack detection,'' \emph{IEEE Access},
  vol.~11, pp. 18\,448--18\,460, 2023.

\bibitem{29}
Z.~Wang, Z.~Zhang, Y.~Tian, Q.~Yang, H.~Shan, W.~Wang, and T.~Q. Quek,
  ``Asynchronous federated learning over wireless communication networks,''
  \emph{IEEE Transactions on Wireless Communications}, vol.~21, no.~9, pp.
  6961--6978, 2022.

\bibitem{30}
H.-S. Lee and J.-W. Lee, ``Adaptive transmission scheduling in wireless
  networks for asynchronous federated learning,'' \emph{IEEE Journal on
  Selected Areas in Communications}, vol.~39, no.~12, pp. 3673--3687, 2021.

\bibitem{44}
Q.~Wu, S.~Wang, P.~Fan, and Q.~Fan, ``Deep reinforcement learning based vehicle
  selection for asynchronous federated learning enabled vehicular edge
  computing,'' in \emph{International Congress on Communications, Networking,
  and Information Systems}.\hskip 1em plus 0.5em minus 0.4em\relax Springer,
  2023, pp. 3--26.

\bibitem{31}
A.~Vedant, A.~Yadav, S.~Sharma, O.~Thite, and A.~Sheikh, ``A practical
  byzantine fault tolerance blockchain for securing vehicle-to-grid energy
  trading,'' in \emph{2022 Global Energy Conference (GEC)}.\hskip 1em plus
  0.5em minus 0.4em\relax IEEE, 2022, pp. 288--293.

\bibitem{32}
X.~Ma, Q.~Jiang, M.~Shojafar, M.~Alazab, S.~Kumar, and S.~Kumari, ``Disbezant:
  secure and robust federated learning against byzantine attack in iot-enabled
  mts,'' \emph{IEEE Transactions on Intelligent Transportation Systems},
  vol.~24, no.~2, pp. 2492--2502, 2022.

\bibitem{33}
A.~Sheikh, V.~Kamuni, A.~Urooj, S.~Wagh, N.~Singh, and D.~Patel, ``Secured
  energy trading using byzantine-based blockchain consensus,'' \emph{IEEE
  Access}, vol.~8, pp. 8554--8571, 2019.

\bibitem{34}
Q.~Wang, T.~Ji, Y.~Guo, L.~Yu, X.~Chen, and P.~Li, ``Trafficchain: A
  blockchain-based secure and privacy-preserving traffic map,'' \emph{IEEE
  Access}, vol.~8, pp. 60\,598--60\,612, 2020.

\bibitem{35}
Y.~Huang, J.~Wang, C.~Jiang, H.~Zhang, and V.~C. Leung, ``Vehicular network
  based reliable traffic density estimation,'' in \emph{2016 IEEE 83rd
  Vehicular Technology Conference (VTC Spring)}.\hskip 1em plus 0.5em minus
  0.4em\relax IEEE, 2016, pp. 1--5.

\bibitem{36}
J.-H. Chen, M.-R. Chen, G.-Q. Zeng, and J.-S. Weng, ``Bdfl: A
  byzantine-fault-tolerance decentralized federated learning method for
  autonomous vehicle,'' \emph{IEEE Transactions on Vehicular Technology},
  vol.~70, no.~9, pp. 8639--8652, 2021.

\bibitem{37}
Q.~Wang, T.~Ji, Y.~Guo, L.~Yu, X.~Chen, and P.~Li, ``Trafficchain: A
  blockchain-based secure and privacy-preserving traffic map,'' \emph{IEEE
  Access}, vol.~8, pp. 60\,598--60\,612, 2020.

\bibitem{38}
J.-w. Xu, K.~Ota, M.-x. Dong, A.-f. Liu, and Q.~Li, ``Siotfog:
  Byzantine-resilient iot fog networking,'' \emph{Frontiers of information
  technology \& electronic engineering}, vol.~19, no.~12, pp. 1546--1557, 2018.

\bibitem{39}
J.~Wang, Y.~Huang, Z.~Feng, C.~Jiang, H.~Zhang, and V.~C. Leung, ``Reliable
  traffic density estimation in vehicular network,'' \emph{IEEE Transactions on
  Vehicular Technology}, vol.~67, no.~7, pp. 6424--6437, 2018.

\bibitem{45}
M.~Fang, J.~Liu, N.~Z. Gong, and E.~S. Bentley, ``Aflguard: Byzantine-robust
  asynchronous federated learning,'' in \emph{Proceedings of the 38th Annual
  Computer Security Applications Conference}, 2022, pp. 632--646.

\bibitem{40}
D.~Long, Q.~Wu, Q.~Fan, P.~Fan, Z.~Li, and J.~Fan, ``A power allocation scheme
  for mimo-noma and d2d vehicular edge computing based on decentralized drl,''
  \emph{Sensors}, vol.~23, no.~7, p. 3449, 2023.

\bibitem{48}
J.~Zhao, H.~Quan, M.~Xia, and D.~Wang, ``Adaptive resource allocation for
  mobile edge computing in internet of vehicles: A deep reinforcement learning
  approach,'' \emph{IEEE Transactions on Vehicular Technology}, 2023.

\bibitem{53}
Q.~Wu, Z.~Zhang, H.~Zhu, P.~Fan, Q.~Fan, H.~Zhu, and J.~Wang, ``Deep
  reinforcement learning-based power allocation for minimizing age of
  information and energy consumption in multi-input multi-output and
  non-orthogonal multiple access internet of things systems,'' \emph{Sensors},
  vol.~23, no.~24, p. 9687, 2023.

\bibitem{54}
Q.~Wu, S.~Shi, Z.~Wan, Q.~Fan, P.~Fan, and C.~Zhang, ``Towards v2i age-aware
  fairness access: A dqn based intelligent vehicular node training and test
  method,'' \emph{Chinese Journal of Electronics}, vol.~32, no.~6, pp.
  1230--1244, 2023.

\bibitem{41}
H.~Q. Ngo, E.~G. Larsson, and T.~L. Marzetta, ``Energy and spectral efficiency
  of very large multiuser mimo systems,'' \emph{IEEE Transactions on
  Communications}, vol.~61, no.~4, pp. 1436--1449, 2013.

\bibitem{49}
S.~Zhang, T.~Cai, D.~Wu, D.~Schupke, N.~Ansari, and C.~Cavdar, ``Iort data
  collection with leo satellite-assisted and cache-enabled uav: A deep
  reinforcement learning approach,'' \emph{IEEE Transactions on Vehicular
  Technology}, 2023.

\bibitem{43}
D.~Silver, G.~Lever, N.~Heess, T.~Degris, D.~Wierstra, and M.~Riedmiller,
  ``Deterministic policy gradient algorithms,'' in \emph{International
  conference on machine learning}.\hskip 1em plus 0.5em minus 0.4em\relax Pmlr,
  2014, pp. 387--395.

\bibitem{55}
T.~P. Lillicrap, J.~J. Hunt, A.~Pritzel, N.~Heess, T.~Erez, Y.~Tassa,
  D.~Silver, and D.~Wierstra, ``Continuous control with deep reinforcement
  learning,'' \emph{arXiv preprint arXiv:1509.02971}, 2015.

\end{thebibliography}
%\biographies

\end{document}